\documentclass{article}

\PassOptionsToPackage{square, numbers, sort&compress}{natbib}

\usepackage[final]{neurips_2023}

\usepackage[utf8]{inputenc} 
\usepackage[T1]{fontenc}    
\usepackage{url}            
\usepackage{booktabs}       
\usepackage{amsfonts}       
\usepackage{nicefrac}       
\usepackage[usenames,dvipsnames]{xcolor}
\usepackage[colorlinks,linktoc=all]{hyperref}
\hypersetup{citecolor=MidnightBlue}
\hypersetup{urlcolor=MidnightBlue}
\hypersetup{linkcolor=black}
\usepackage{multirow, tabularx}
\usepackage{comment}
\usepackage{enumitem}
\setlist[itemize]{leftmargin=5mm}
\setlist[enumerate]{leftmargin=8mm}
\usepackage{subcaption}
\usepackage{pgfplots}
\usepackage{upgreek}
\pgfplotsset{compat=1.16}
\usepackage{siunitx}   
\usepackage{wrapfig}
\usepackage{MnSymbol}
\usepackage{algorithm}
\usepackage{algpseudocode}
\usepackage{setspace}
\usepackage[draft]{minted}
\newcommand\Algphase[1]{%
\vspace*{-.7\baselineskip}\Statex\hspace*{\dimexpr-\algorithmicindent-2pt\relax}\rule{\textwidth}{0.4pt}%
\Statex\hspace*{-\algorithmicindent}\textbf{#1}%
\vspace*{-.7\baselineskip}\Statex\hspace*{\dimexpr-\algorithmicindent-2pt\relax}\rule{\textwidth}{0.4pt}%
}

\algnewcommand\algorithmicassume{\textbf{Assume:}}
\algnewcommand\Assume{\item[\algorithmicassume]}

\usepackage{amsthm}

\makeatletter
\newtheorem*{rep@theorem}{\rep@title}
\newcommand{\newreptheorem}[2]{%
\newenvironment{rep#1}[1]{%
 \def\rep@title{#2 \ref{##1}}%
 \begin{rep@theorem}}%
 {\end{rep@theorem}}}
\makeatother

\newreptheorem{definition}{Definition}
\newreptheorem{lemma}{Lemma}

\usepackage{tikz}
\usetikzlibrary{bayesnet}
\usetikzlibrary{arrows}
\usetikzlibrary{patterns}
\usepackage{color}
\usepackage{caption}

\usepackage{amsthm}

\newcommand{\auxdata}{auxiliary data}
\newcommand{\ours}{\textit{CATO} }

\newtheorem{lemma}{Lemma}
\newtheorem{definition}{Definition}
\newtheorem{assumption}{Assumption}

\makeatletter
\newcommand*{\indep}{%
  \mathbin{%
    \mathpalette{\@indep}{}%
  }%
}
\newcommand*{\nindep}{%
  \mathbin{
    \mathpalette{\@indep}{\not}
  }%
}
\newcommand*{\@indep}[2]{%
  \sbox0{$#1\perp\m@th$}
  \sbox2{$#1=$}
  \sbox4{$#1\vcenter{}$}
  \rlap{\copy0}
  \dimen@=\dimexpr\ht2-\ht4-.2pt\relax
  \kern\dimen@
  {#2}%
  \kern\dimen@
  \copy0 
} 
\makeatother

\usepackage{amsmath,amsfonts,bm}

\def\eqref#1{equation~\ref{#1}}

\def\1{\bm{1}}

\def\rvm{{\mathbf{m}}}

\def\rvs{{\mathbf{s}}}

\def\rvw{{\mathbf{w}}}
\def\rvx{{\mathbf{x}}}

\def\vmu{{\bm{\mu}}}

\def\vrho{{\bm{\rho}}}

\DeclareMathAlphabet{\mathsfit}{\encodingdefault}{\sfdefault}{m}{sl}
\SetMathAlphabet{\mathsfit}{bold}{\encodingdefault}{\sfdefault}{bx}{n}

\def\gH{{\mathcal{H}}}

\def\gM{{\mathcal{M}}}
\def\gN{{\mathcal{N}}}

\def\gP{{\mathcal{P}}}

\def\gR{{\mathcal{R}}}

\def\gX{{\mathcal{X}}}
\def\gY{{\mathcal{Y}}}

\newcommand{\E}{\mathbb{E}}

\newcommand{\R}{\mathbb{R}}

\usepackage{cleveref}
\usepackage{pgfplots}
\pgfplotsset{compat=1.18}

\title{Data Augmentations for Improved (Large) Language Model Generalization}

\author{%
  Amir Feder \thanks{Equal Contribution. Correspondence to 
  \href{mailto:amir.feder@columbia.edu?subject=[CausalAug] ...}{\texttt{amir.feder@columbia.edu}}} $^{\;1,2}$, Yoav Wald $^{*\;3}$, Claudia Shi $^{1}$, Suchi Saria $^{3}$ and David Blei $^{1}$ \vspace{1mm}\\
  $^{1}$ Columbia University, $^{2}$ Google Research, $^{3}$ Johns Hopkins University
}

\begin{document}

\maketitle

\begin{abstract}
The reliance of text classifiers on spurious correlations can lead to poor generalization at deployment, raising concerns about their use in safety-critical domains such as healthcare. In this work, we propose to use counterfactual data augmentation, guided by knowledge of the causal structure of the data, to simulate interventions on spurious features and to learn more robust text classifiers. We show that this strategy is appropriate in prediction problems where the label is spuriously correlated with an attribute. Under the assumptions of such problems, we discuss the favorable sample complexity of counterfactual data augmentation, compared to importance re-weighting. Pragmatically, we match examples using auxiliary data, based on diff-in-diff methodology, and use a large language model (LLM) to represent a conditional probability of text. Through extensive experimentation on learning caregiver-invariant predictors of clinical diagnoses from medical narratives and on semi-synthetic data, we demonstrate that our method for simulating interventions improves out-of-distribution (OOD) accuracy compared to baseline invariant learning algorithms.
\end{abstract}

\section{Introduction}

The reliance on spurious correlations is a significant challenge for Machine Learning (ML) safety as it can lead to performance degradation of deployed models. Spurious correlations are prevalent in various applications such as medical imaging \cite{zech2018variable, degrave2021ai}, text classification \cite{mccoy2019right}, and risk prediction systems \cite{caruana2015intelligible}. 
Failures due to spurious correlations occur under distribution shift \citep{quinonero2008dataset, subbaswamy2019preventing, finlayson2021clinician}, which may result from differences in data recording protocols, shifts in the underlying population being monitored, or the way the ML tool is being used. 
In this paper, we focus on text classification and explore how using language models in a domain-informed way can help us avoid reliance on spurious correlations.

Consider a scenario where we want to make robust predictions about patients' conditions, probability of readmission, etc., using clinical narratives written in hospitals \cite{spyns1996natural, zhou2007temporal, wu2020deep}. In this setting, a common issue arises due to clinical practice, where patients with certain conditions are directed to specific caregivers in the hospital. 
When we train a predictor from a single dataset that exhibits some correlation between caregiver-specific style and clinical outcomes, the predictor may unintentionally rely on the style to make predictions. This leads to poor generalization on unseen hospitals, i.e. failure to generalize out of distribution(OOD), due to changes in clinical practice \citep{finlayson2021clinician}. However, collecting a dataset that is large enough to avoid such spurious associations is infeasible due to various reasons such as rare conditions, privacy concerns, etc. To tackle this problem, we propose leveraging available \auxdata{} (e.g., time, document type, demographics) and incorporating knowledge about the causal structure of the problem to build a more robust classifier. For example, in the note classification task, we can use our knowledge that some \auxdata{}, such as the patient's current state, can affect doctor assignment, to improve the classifier's robustness.

Causal inference often makes use of such \auxdata{} and has now been used in a variety of ways to improve OOD generalization \citep{subbaswamy2019preventing, peters2016causal, magliacane2018domain, arjovsky2019invariant, subbaswamy2022unifying}.
Data augmentation methods have demonstrated impressive performance in these tasks as well \citep{robey2021modelbased, yao2022improving, gao2023out}, and with recent improvements in generative models, forming additional principles to incorporate domain knowledge into data augmentations seems like a promising path forward. 

In this work we pursue this and develop \emph{causally-driven data augmentation methods}, that leverage \auxdata{} and domain knowledge. Intuitively, generating versions of clinical narratives as if they had been written by different caregivers, de-correlates the writing style from the patient condition we wish to predict. However, such data generation can be difficult to achieve in practice and problem-specific traits must be taken into account \citep{kocaoglu2018causalgan}. Observing that data augmentation can be treated as counterfactual outcome estimation under a causal formalism, motivates the use of causal inference methods that are commonly used for such tasks across the sciences. While our approach can be applied to many modalities of data, in this work we focus on text classification and harness the recent advances in LLMs towards counterfactual estimation. Our contributions are:
\begin{enumerate}
    \item Through extensive experiments, we show how the use of language models in a manner that is informed by causal knowledge improves model robustness in challenging safety-critical tasks in healthcare. Furthermore, our findings are reinforced by experiments that incorporate semi-synthetic scenarios, and simulations where there are ground-truth counterfactuals. 
    \item We formalize counterfactual data augmentation in a prediction setting as a method to deconfound the target and a spuriously correlated attribute. We show how deconfounding improves OOD generalization. In a setting where sample complexities for alternative methods (re-weighting and invariance penalties) can be derived, we show favorable generalization bounds for accurately performed data-augmentation.
    \item Our data-augmentation methods rely on common assumptions in the causal inference literature such as no unmeasured confounding and parallel trends in diff-in-diff \citep{abadie2005semiparametric}, applied with LLMs. We believe that leveraging \auxdata{} and assumptions about causal structure, along with the use of LLMs and other generative models, can be a fruitful framework for addressing many out-of-distribution generalization problems.
\end{enumerate}

Next, we provide a brief survey of relevant work (\S\ref{sec:rel}). We then present a formal setting motivating counterfactual augmentation for OOD generalization (\S\ref{sec:setting}), our methods for counterfactual estimation and reason formally about the preferable sample complexity of our approach  (\S\ref{sec:cda}). Finally, we present our main experimental results (\S\ref{sec:experiments}) and discuss limitations and future directions (\S\ref{sec:dis}).

\section{Related Work}
\label{sec:rel}

\textbf{Invariant and Shift-stable Learning.}
This paper contributes to the growing literature on invariant and shift-stable learning, which tackles the problem of learning models that generalizes across different distributions or settings. Invariant learning through feature pruning was pioneered by \citet{peters2016causal}, and has since been developed for variable selection \citep{magliacane2018domain, heinze2018invariant} and representation learning \citep{li2018deep, arjovsky2019invariant, wald2021calibration, krueger2021out, puli2022outofdistribution, makar2022causally, jiang2022invariant}. These methods have been applied in a range of domains, including natural science \citep{peters2016causal, magliacane2018domain, heinze2018invariant}, causal estimation \citep{shi2021invariant, yin2021optimization}, computer vision \citep{arjovsky2019invariant, krueger2021out}, and NLP \citep{veitch2021counterfactual, dranker2021irm, feder2021causal, feder2022eye}. 
However, recent studies have highlighted limitations in many invariant learning approaches, particularly in achieving conditional independence \citep{kamath2021does, rosenfeld2020risks, guo2021out, wald2022malign}.
Others have investigated learning of stable models by leveraging causal methods through techniques like graph-surgery \cite{subbaswamy2019preventing, subbaswamy2022unifying}, that come with generalization guarantees. Yet others have explored the advantages of data augmentation \cite{kaushik2019learning, kaushik2020explaining}. In this work, we combine the latter two approaches to improve OOD generalization for text based classification. 

\textbf{Counterfactually Augmented Data.}
To learn invariant predictors, a popular and straightforward approach is data augmentation. When data augmentation involves actions that go beyond simple manipulations (e.g. image rotations, crops etc.), it is often referred to as \emph{counterfactual data augmentation} \citep{kaushik2019learning}. Constructing counterfactual instances that involve perturbations to confounding factors \cite{garg2019counterfactual}, or to the label \cite{kaushik2019learning,kaushik2020explaining, jha2020does}, and incorporating them into the training data, breaks up correlations that we do not wish our model to exploit towards prediction. Most work on counterfactual data augmentation in text involve manual editing by humans, heuristic keyword replacement, or automated text rewriting \cite{kaushik2019learning, gardner-etal-2020-evaluating, shekhar-etal-2017-foil, garg2019counterfactual, feder2021causalm, zmigrod-etal-2019-counterfactual, riley2020textsettr, wu2021polyjuice, mao2021generative, rosenberg2021vqa, abraham2022cebab, wu2023causal}. Manual editing is accurate and effective \citep{kaushik2020explaining, joshi2022investigation} but expensive, hence our goal is to \emph{make counterfactual data augmentation scalable}, demanding smaller human effort. Keyword-based methods can be limited in coverage and difficult to generalize across languages \cite{antoniak-mimno-2021-bad}. Generative approaches offer a balance of fluency and coverage \cite{zhou2023implicit}, but generating meaningful counterfactuals is challenging \cite{Calderon:22}. Our work departs from previous techniques by using \emph{causal \auxdata{} structure and LLMs} to alleviate this challenge and generate plausible counterfactual data augmentations.

\textbf{Clinical Notes.}
Clinical notes are the backbone of electronic health records, often containing vital information not observed in other structured data \citet{kreimeyer2017natural}. 
Clinical NLP involves identifying this information, and standardized datasets and competitions exist for this purpose \citep{uzuner2009recognizing,savova2010mayo,jensen2012mining,ford2016extracting,zhu2018clinical}. Best performing approaches have leveraged transformer architectures both for token-level classification tasks \citep{peng2019transfer,yadav2019survey,si2019enhancing,lee2020biobert}, and for using complete clinical records \cite{roussinov2022predicting, seinen2022use}. 
Recently, large language models (LLMs), similar to those we use to generate counterfactual notes, were shown to have clear potential for improving clinical NLP systems \cite{singhal2022large, ayers2023comparing}.
In our experiments, we follow recent papers in clinical NLP addressing challenges of degraded performance across different hospitals \citep{feder2022building,zhang2022section,feder2020active}.

\section{Problem Setting} \label{sec:setting}
To formally analyze how counterfactual data augmentation helps OOD generalization, we consider a setting where the label is spuriously correlated with a known attribute. This setting has been used previously to study learning with ``shortcuts" \citep{makar2022causally} and  spurious correlations \citep{veitch2021counterfactual}. We note that our approach is applicable and valid under additional settings and causal graphs (e.g. “purely spurious” problems defined in \citet{wang2022unified}) and we elaborate on this at \cref{app:causal_structure}. The data generating process used here motivates counterfactual data augmentation in a principled manner, as it describes the main problem we study and it is possible to analytically compare sample complexity with an alternative solution (see \cref{sec:alg_analysis}).
\begin{wrapfigure}{r}{0pt}
        \centering 
        \raisebox{0pt}[\dimexpr\height-0.6\baselineskip\relax]{
        \includegraphics[scale=0.8]{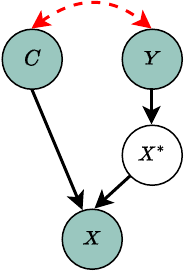}}%
        \caption{Prediction problem with a spuriously correlated attribute.}
        \label{fig:dgp_purely_spu_ac}
\end{wrapfigure}

Consider a classification problem with $L$ classes, where the label $Y$ is correlated with a certain attribute $C$ in the training data and this correlation may change arbitrarily at test time (denoted by a red edge $C {\color{red}\leftrightarrow} Y$ in \cref{fig:dgp_purely_spu_ac}). In our medical notes example, $C$ is the caregiver writing the note and $Y$ is the underlying condition we wish to diagnose. We denote the number of caregivers in our training data by $[K]$. For a given loss function $\ell:\R^{L}\times [L] \rightarrow \R$ and distribution $P$, we denote the expected loss of a hypothesis $h:\gX\rightarrow \R^L$ by $\gR^{\ell}_{P}(h)$ and its expected accuracy by $\gR^{\ell_{01}}_{P}(h)$. The data-generating process is depicted by the causal model in \cref{fig:dgp_purely_spu_ac}, for our motivating example of clinical notes classification $X$ is a vector representation of the clinical note and $X^*$ is an unobserved sufficient statistic, representing all the relevant information about $Y$ in the note that is unaffected by the writing style of the caregiver. Let us formally define this setting. 
\begin{definition} \label{def:prob_setting}
We denote the set of distributions induced by interventions on a causal model with the structure in \cref{fig:dgp_purely_spu_ac} by
\begin{align*}
\gP = \left\{ P(X \mid X^*, C)P(X^* \mid Y)P(Y)\tilde{P}(C \mid Y) ~:~ \tilde{P}(C \mid Y=y)\in\Delta^{K-1} ~~ \forall y\in{[L]} \right\},
\end{align*}
where all distributions other than $\tilde{P}(C \mid Y)$ are fixed. In a prediction problem with a spuriously correlated attribute, the learner is provided with a set $\left\{ (\rvx_i, y_i, c_i) \right\}_{i=1}^{N}$ sampled i.i.d from $P_{\text{train}}\in{\gP}$. We assume that $X^* = e(X)$ almost surely for some $e:\R^d \rightarrow \R^{d^*}$.
\end{definition}
In this problem, once $X^*$ is recovered no additional information from $X$ is needed to predict $Y$. We can also see from the graph that interventions on $\tilde{P}(C \mid Y)$ do not change the conditional distribution $P(Y \mid X^*)$. Therefore an optimal solution that does not rely on $C$ is $h^*(\rvx) = \mathrm{arg}\max_{y\in{L}}{P(Y=y \mid e(\rvx))}$.
In clinical note classification, $X^*$ represents all the information in the note about the patient conditions, unsullied by the writing style of caretaker $C$. To obtain $h^*(\rvx)$ we will rely on risk minimization w.r.t a distribution where $Y$ and $C$ are uncorrelated.

\subsection{Learning Robust Classifiers when Counterfactuals are Available}
Consider the unconfounded distribution $P_\bot\in{\gP}$ that is given by intervening on $C$, setting it independent of $Y$ and uniformly distributed, $\tilde{P}(C \mid Y) = P_{\text{unif}}(C)$. An optimal classifier under $P_{\bot}$ has the following min-max optimality guarantee. \footnote{This claim is shown in \citet{makar2022causally}, \cref{app:formal_things} includes a proof for completeness. We set the distribution over $C$ in $P_\bot$ as uniform for simplicity, the derivation for non-uniform distributions is analogous.}

\begin{lemma} \label{lem:unconfounded_opt}
For the prediction problem in \cref{def:prob_setting}, the Bayes optimal classifier under the unconfounded distribution $P_\bot\in{\gP}$ where $C$ is uniformly distributed and independent of $Y$ is $h^*(\rvx) = \mathrm{arg}\max_{y\in{[K]}} P_\bot(Y=y \mid X^*=e(\rvx))$. It is a minimizer of $\min_{h:\gX\rightarrow [L] }\max_{P\in{\gP}}{\gR^{\ell_{01}}_{P}(h)}$ and $\gR^{\ell_{01}}_{P}(h^*)=\gR^{\ell_{01}}_{P_{\bot}}(h^*)$ for all $P\in{\gP}$.
\end{lemma}
Hence we would like to minimize risk w.r.t $P_{\bot}$ and we cannot do that directly by via ERM since our training data is sampled from $P_{\text{train}} \neq P_\bot$.
Instead we consider risk minimization over an augmented dataset that contains counterfactual instantiations of our training data under different values of $C$.

\textbf{Minimizing $\gR_{P_{\bot}}$ via Counterfactual Data Augmentation.} Returning to our motivating example, assume that we could generate clinical notes for all alternative scenarios. That is, obtain the clinical notes that would have been written if each patient had been seen by all possible caregivers $c\in{[K]}$ and each caregiver had written their own version of the note $\rvx_i(c)$.
Given these counterfactual clinical notes, we seek a hypothesis that minimizes the average loss over all such possible scenarios, denoted by $\widehat{\gR}^{\ell}_{\text{aug}}(h)$.
\begin{definition}
Consider a prediction problem with a spuriously-correlated attribute (see \Cref{def:prob_setting}). For a given example $\rvx_i$, we denote its counterfactual with attribute value $c\in{[K]}$ as derived from the corresponding causal model, by $\rvx_i(c)$. For estimates of the counterfactuals $\left\{ \hat{\rvx}_i(c) \right\}_{i\in{[N]}, c\in{[K]}}$ and a hypothesis $h\in\gH$, the counterfactually augmented empirical risk is
\begin{align} \label{eq:aug_erm}
    \widehat{\gR}^{\ell}_{\text{aug}}(h) = \frac{1}{NK}\sum_{i\in{[N]}, c\in{[K]}}{\ell \left( h\left(\hat{\rvx}_i(c)\right), y_i \right)}.
\end{align}
\end{definition}
We use approximate counterfactuals $\hat{\rvx}_i(c)$ in our definition to highlight that in practice we cannot obtain a precise estimate of $\rvx_i(c)$. In the ideal case where $\hat{\rvx}_i(x) = \rvx_i(c)$, the expected loss $\gR^{\ell}_{\text{aug}}(h)$ where $N\rightarrow \infty$, satisfies $\gR^{\ell}_{\text{aug}}(h) = \gR^{\ell}_{P_{\bot}}(h)$. This follows by a simple derivation and it is part of a claim we give later in \Cref{lem:generalization_bound}. Hence obtaining this dataset is useful for our goal of minimizing risk under $P_{\bot}$. Our main challenge is then to derive effective approximations for counterfactuals such as clinical notes under alternative writing styles.
\section{Assumptions and Algorithms for Estimating Counterfactuals} \label{sec:cda}
Perfectly capturing writing style is a strong assumption. Even if we could perfectly model writing styles, we only observe a limited set of variables - the actual notes $x$, outcomes $y$, and assigned caregivers $c$. We do not observe all factors that could influence what each caregiver would write.
To alleviate this problem, we make use of \auxdata{} $M$ that is available during training, but might not be available in deployment.

As an example, consider two caregivers $c$ and $\tilde{c}$, where a note $\rvx_i$ was  written by $c_i = \tilde{c}$. We want to estimate what $\rvx_i(c)$, the note caregiver $c$ would have written, might look like. To this end we will build a model $\tau_c(\cdot)$ that takes data and generates a note in caregiver $c$'s style.
Now suppose caregiver $c$ usually sees patients with high blood pressure and always includes blood pressure values in notes, while $\tilde{c}$ rarely does. A naive model estimating $\hat{\rvx}_i(c) = \tau_c(\rvx_i)$ based only on $c$'s notes may fill in false blood pressure information, conflating that with $c$'s style.
Including vitals data like blood pressure, typically recorded in a patient's health record, can provide additional context for our model. This extra information can assist the model in reasoning about external/background variables, leading to more accurate estimates.

\subsection{Identification of the Counterfactual Distributions}

To make effective use of this data, we suggest that the input to the model $\tau_c:\gX\times\gM\rightarrow \gX$ will include a baseline text to be edited and \auxdata{} $\rvm$. Intuitively, accounting for confounding between the identity of the caregiver $C$ and the text $X$, with auxiliary data $M$ should result in improved augmentation. 

We formalize this intuition using an assumption from causal inference. To identify the counterfactual text distributions using the observed distribution, we assume strong ignorability \citep{imbens2009recent, pearl2009causality, shalit2017estimating}
\begin{assumption}[Strong ignorability] \label{ass:strong_ignorability}
For all $P\in{\mathcal{P}}$ it holds that $X(c) \indep C \mid M$, and for all values of $\rvm\in{\gM}$, $P(\rvm) > 0$.
\end{assumption}
Under this assumption, we can rewrite the counterfactual distribution with the observed distribution,
\begin{align*}
P(X(c)) = \int P(X(c) \mid M= \rvm)P(M=\rvm) d\rvm = \int P(X \mid C=c, M= \rvm)P(M=\rvm) d\rvm.
\end{align*}

However, in practice, we do not observe many samples from $ P(X \mid C=c, M= \rvm)$, making it a poor approximation for the counterfactual distribution.
We address this by using counterfactual data augmentation \citep{kaushik2019learning}.
Formally, we assume that for all possible counterfactual distributions $c \in [K]$, there exist a function $\tau_c$ that maps from the observed distribution $P(X \mid M=\rvm)$ to the target counterfactual distribution $P(X(c) \mid M=m)$.

We approximate the loss under the counterfactual distributions through the empirical loss produced by data augmentation. That is, for a hypothesis $h\in{\gH}$
\begin{align*}
\E_{P(X(c))}[\ell(h(\rvx), y)] \approx \frac{1}{N} \sum_{i\in{[N]}} \tau_{c}(\rvx_i, \rvm_i).
\end{align*}
Note that whenever the text in the training set is already written by caregiver $c$, i.e. $c_i=c$, we will simply keep the original text $\rvx_i$

\paragraph{Evaluation of Augmented Distribution.}
The right hand-side of the above equation is a Monte-Carlo estimator of the distribution of augmented notes, which averages the distributions $\tau_{*, c}(P_{\text{train}}(X, M))$ over all caregivers $c\in{[K]}$. The distribution $\tau_{*, c}(P_{\text{train}}(X, M))$ is aimed to follow the style of caregiver $c$.
While the observed samples from one counterfactual distribution may not be sufficient to approximate the whole distribution, they can be used to assess the quality of the counterfactual augmentation algorithm $\tau_c$.

High-quality counterfactual estimation, as measured by small distributional divergence between our estimator and the target distribution, will help in lowering the upper bound on the risk $\gR^{\ell}_{P_{\bot}}(h)$ (see \cref{lem:generalization_bound} in \cref{sec:alg_analysis}).
Then to estimate divergences between these two distributions, we may use validation sets from our training data. A sample from $\tau_{*, c}(P_{\text{train}}(X, M))$ is obtained simply by running training data through $\tau_c$, while a sample from $P(X(c))$ can be obtained either by adjusting for $M$, or we can obtain a sample from $P(X \mid C=c, M= \rvm)$ for each value of $\rvm$ and compare that to a sample obtained by augmenting validation data where $M=\rvm$. In both cases two-sample tests can be applied and obtain estimates of divergences between the two distributions. That is of course as long as positivity holds, i.e. the second part of the assumption, as otherwise we will not be able to obtain samples of $P(X \mid C=c, M= \rvm)$ for certain values of $\rvm$ and $c$.

We now describe the estimation methods that obtain $\tau_c$. The methods are based on classical causal inference methods, applied to our high-dimensional setting, and relying on the \auxdata{} $M$.

\subsection{Methods for Estimation of Counterfactuals}
\begin{figure}
  \begin{minipage}[t][6cm][t]{0.46\textwidth}
  \vspace{-19pt}
    \begin{algorithm}[H]
    \caption{\ours}
    \label{alg:cdaug}
    \begin{algorithmic}[1]
        \Require Training set $\{(\rvx_i, y_i, c_i, \rvm_i)\}_{i=1}^{N}$ \newline
        Hypothesis class $\gH$
        \newline
        $\mathrm{Version}\in{\left\{(A), (B)\right\}}$ \newline
        {\bfseries Optional} pre-treatment data $\{(\rvx_{\text{pre}, i})\}_{i=1}^{N}$
        \Ensure A hypothesis $h_{\text{aug}}(\rvx)$ \If{$\mathrm{Version} = (A)$}
        \State Get $\tau_c(\rvm, \rvx)$ with preprocess (A)
        \State Get $\hat{\rvx}_i(c) = \tau_c(\rvx_{i, \text{pre}}, \rvm_i) ~ \forall i\in{[N]}$
        \Else
        \State Get $\tau_c(\rvm, \rvx)$ with preprocess (B)
        \State Get $\hat{\rvx}_i(c) = \tau_c(\rvx_i, \rvm_i) ~ \forall i\in{[N]}$
        \EndIf \\
        \Return $h_{\text{aug}}\in{\gH}$ that minimizes $\widehat{\gR}^{\ell}_{\text{aug}}$.
        \Statex
        \vspace{-3pt}
    \end{algorithmic}
    \end{algorithm}
  \end{minipage}
  \hspace{0.02\linewidth}
\begin{minipage}[t][5cm][t]{0.47\textwidth}
    \begin{algorithmic}[1]
        \setcounter{ALG@line}{0}
        \Algphase{Pre-process \ours (A)}
        \Assume $\rvm$ includes the label $y$ and pre-treatment attribute $c_{\text{pre}}$, among other \auxdata. We are given $\{\rvx_{j, \text{pre}}\}_{j=1}^{N}$. \vspace{2pt}
        \State Set $\vrho(c_j, \rvm_j) = \rvx_j - \rvx_{j, \text{pre}}$ for $j\in{[N]}$. \\
        \Return $\tau_c(\rvx, \rvm) := \rvx_{ \text{pre}} + \rho(c, \rvm)$
        \vspace{2pt}
        \vspace{8pt}
        \setcounter{ALG@line}{0}
        \Algphase{Pre-process \ours (B)}
        
        \Assume $\rvm$ includes the label $y$ among other \auxdata. \\
        \vspace{2pt}        
        \Return prompt $\tau_c(\rvx, \rvm)$ that rewrites $\rvx$ in the style of matching examples with attribute $c$, i.e. $\{\rvx_j : (\rvm_j, c_j) = (\rvm, c) \}$.
    \end{algorithmic}
    \hrulefill
\end{minipage}
\vspace{-5mm}
\end{figure}

Counterfactual estimation is an established problem in causal effect estimation \cite{rosenbaum1983central, pearl2009causality, imbens2015causal}. Here we adapt identification strategies and estimation procedures in the causal literature to estimate $\rvx_i(c)$. Our framework for estimating counterfactuals \ours 
(\textbf{C}ausal-structure Driven \textbf{A}ugmentations for \textbf{T}ext \textbf{O}OD Generalization)
involves the use of an LLM to model the conditional probability distribution of text. Counterfactuals are formed by matching similar \auxdata{} examples or manipulating texts' vector representations, as described below.

\textbf{Prompting with matched examples.} Our first estimation method in \Cref{alg:cdaug}(B) draws insights from matching \citep{rosenbaum1983central}. We construct a prompt for an LLM, that given an original text $\rvx$ and a set of context notes, asks the LLM to rewrite $\rvx$ in their style. Now given text $\rvx$ with \auxdata{} $\rvm$ that we wish to estimate with counterfactual value $c$ (i.e. writing style), $\tau_c(\rvx, \rvm)$ runs this prompt with context notes whose \auxdata{} is similar to $\rvm$ and their attribute value equals the desired $c$.

\textbf{Diff-in-diff estimation.} The procedure we use for medical note generation relies on additional structure involving panel data (i.e. data collected over time intervals across several individuals). In our case of clinical narratives, a narrative is usually consisted of several notes taken over the course of a patient's visit and each may be written by a different caregiver. Prediction is made using the release note from the hospital whose embedding consists our features $\rvx$. For simplicity let us consider a single note $\rvx_{\text{pre}}$ taken prior to $\rvx$.
Difference-in-difference \citep{card1993minimum, abadie2005semiparametric, angrist2009mostly} estimation of causal effect is based on the parallel-trends, or constant effect assumption that two units $i,j$ with similar pre-treatment conditions would have seen the same effect had they been given the same treatment. In our case, the treatment is an assignment to a certain caregiver. Hence we assume our \auxdata{} $\rvm$ includes $c_{\text{pre}}$, the caregiver assigned pre-treatment.
\begin{assumption}[constant effect]
Let $\rvx_{i, \text{pre}}$ be the pre-treatment features for unit $i$, and assume $\rvm_i$ includes the pre-treatment attribute $c_{i, \text{pre}}$.
There exists a function $\vrho: [K] \times \gM \rightarrow \gX$ such that $\rvx_i(c) = \rvx_{i, \text{pre}} + \vrho(c, \rvm_{i})$.
\end{assumption}
\begin{wrapfigure}{r}{0.4\textwidth}
  \label{fig:dind}
  \vspace{-5mm}
  \includegraphics[width=0.4\textwidth]{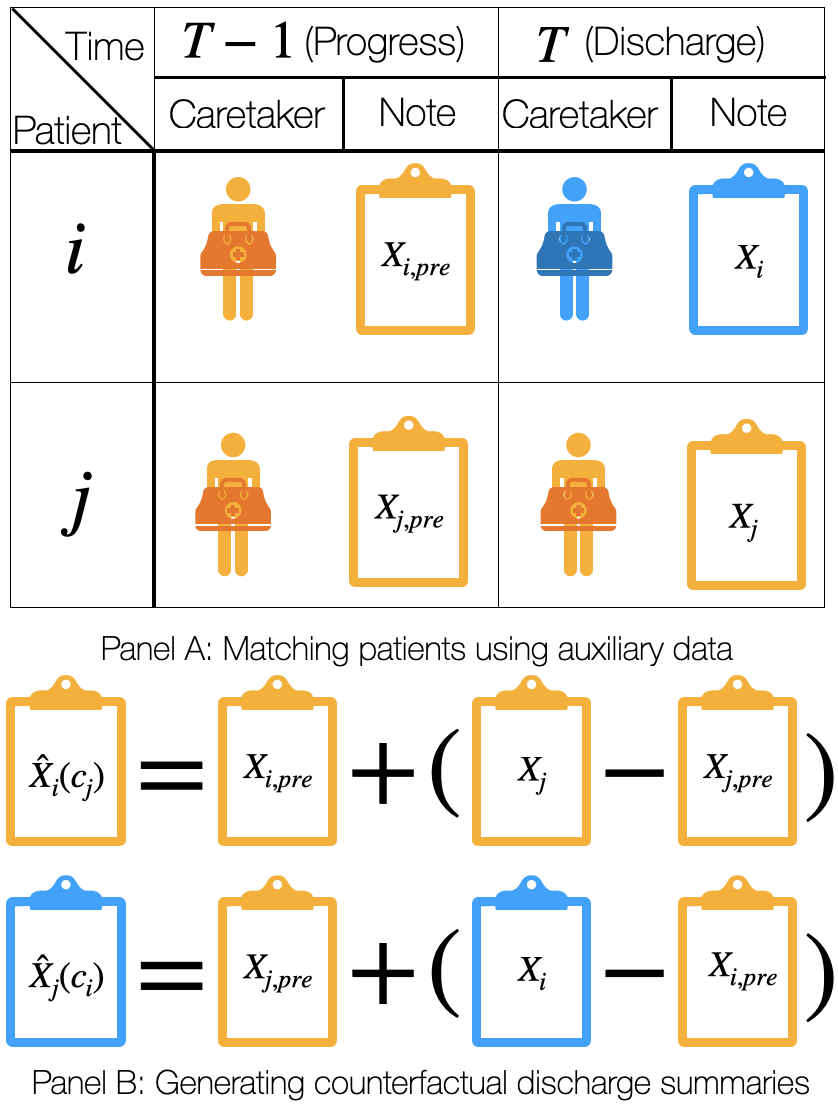}
  \caption{Generating counterfactual clinical notes for patients using \auxdata{} with \Cref{alg:cdaug}(A).}
  \vspace{-5mm}
\end{wrapfigure}
Under this assumption, to calculate $\rvx_{i}(c)$ we can use any unit $j$ for which $\rvm_i=\rvm_j$ and has $c_j=c$ to estimate $\vrho(c, \rvm_i) = \rvx_j - \rvx_{\text{pre}, j}$. The resulting estimation procedure is given in \cref{alg:cdaug}(B) and illustrated in \cref{fig:dind}.

Before empirically evaluating our methods, we discuss alternatives for learning robust classifiers in our setting, and how their properties fair compared to counterfactual augmentation.

\subsection{Why Bother with Counterfactual Data Augmentation?} \label{sec:alg_analysis}

Reasoning about counterfactuals with problem-specific domain knowledge is a considerable challenge, and it is interesting to see whether this has any advantage in learning robust classifiers compared to methods that rely on less stringent assumptions. A simple alternative to approximating counterfactuals involves re-weighting the loss function (see e.g. \citet{shimodaira2000improving, makar2022causally}).

\textbf{Reweighting baseline.}
Intuitively, re-weighting samples from the uncorrelated distribution $P(Y, C) = P(Y)P(C)$ by setting for each example $i$ a weight $w_i = P_{\text{train}}(Y=y_i)P_{\text{train}}(C=c_i) / P_{\text{train}}(Y=y_i, C=c_i)$ and minimizing the weighted empirical risk:
\begin{align*}
    \hat{\gR}^{\ell}_{\rvw}(h) = \frac{1}{m}\sum_{i\in{[m]}}{w_i \ell\left(h(\rvx_i), y_i\right)}.
\end{align*}
It can be proved that at the limit of infinite data the method learns a min-max optimal hypothesis, as it also effectively  minimizes $\gR^{l}_{P_\bot}$  (see \citep{makar2022causally}). 
While augmentations may not seem advantageous for identifying the correct hypothesis, reweighting can demand a larger sample to identify the correct hypothesis, particularly when $Y$ and $C$ are highly correlated.\footnote{We remark that other works discuss the potential benefits of data augmentation for identification in other problem settings, e.g. \citep[Thm.~9]{wang2022unified} and \citep{gao2023out}.}

\textbf{Comparing sample complexities.}
To make this statement precise, we can apply the bounds from \citet{NIPS2010_59c33016} and compare them with an upper bound that we will derive for our method in \Cref{lem:generalization_bound}. To this end, let us consider the exponent of the R\'enyi divergence as a measure of dependence between $Y$ and $C$ in the training data. The divergence is given by $d_{\alpha, \text{train}}\left( Y, C \right) = [\sum_{y\in{[L]}, c\in{[K]}}{P^{\alpha}_{\text{train}}(Y=y, C=c)/P^{\alpha-1}_{\text{train}}(Y=y)P^{\alpha-1}_{\text{train}}(C=c)}]^{\frac{1}{\alpha-1}}$, and we may derive the following bound for a hypothesis $h\in{\gH}$ and any $\delta\in{[0,1]}$:
\begin{align} \label{eq:renyi_bound}
     \gR^{\ell}_{P_\bot}(h) &\leq \widehat{\gR}_{\rvw}^{\ell}(h) + \sqrt{\frac{2 d_{2, \text{train}}\left( Y, C \right) \cdot \log(1 / \delta)}{N}} + \frac{d_{\infty, \text{train}}(Y, C)\cdot \log(1 / \delta)}{N}.
\end{align}
A complementary lower bound on $\widehat{\gR}_{\rvw}^{l}(h)$ can also be derived based on results in \citet{NIPS2010_59c33016}.
To compare this with counterfactual augmentations, denote our augmentation model by $\tau:\gX \times \gM \rightarrow \gX^K$, which is some measurable function whose output's $c$-th coordinate is the counterfactual estimate w.r.t. caregiver $c$, i.e. $\hat{\rvx}(c) = \tau_c(\rvx, \rvm)$. The following statement quantifies the relation between the accuracy of $\tau(\cdot)$ in approximating counterfactuals and the classification accuracy of a model learned from the augmented data, via minimization of $\widehat{\gR}^{\ell}_{\text{aug}}(h)$ in \cref{eq:aug_erm}.
\begin{lemma} \label{lem:generalization_bound}
Consider a prediction problem with a spuriously-correlated attribute (\cref{def:prob_setting}), a measurable function $\tau:\gX\times \gM\rightarrow \gX^K$, and let $d_1(P, Q)$
denote the total variation distance between two distributions $P,Q$. Further let $h^*, h^*_{\text{aug}}$ denote the optimal hypotheses w.r.t $\gR^{\ell_{01}}_{P_\bot}, \gR^{\ell_{01}}_{\text{aug}}$ respectively and let $\lambda_{\text{aug}} = \left[R_{P_\bot}^{\ell_{01}}(h^*_{\text{aug}}) - R_{P_\bot}^{\ell_{01}}(h^*) \right]$. For any hypothesis $h\in\gH$, and any $\delta\in{(0,1)}$ it holds that with probability at least $1-\delta$ over the draw of the training set,
\begin{align*}
    \gR^{\ell_{01}}_{P_\bot}(h) &\leq \widehat{\gR}^{\ell_{01}}_{\text{aug}}(h) + \sqrt{\frac{\log(1 / \delta)}{N}} + K^{-1}\cdot\sum_{c\in{[K]}}d_1\left(\tau_{c,*}\left( P_{\text{train}}\left(X, M\right) \right), P\left( X(c) \right) \right) + \lambda_{\text{aug}}.
\end{align*}
\end{lemma}
The divergence $d_1(\tau_{c, *}(P_{\text{train}}(X, M)), P(X(c)))$ is a distance between the true distribution over counterfactual instances $P(X(c))$ and our augmented data $\tau_{c, *}(P_{\text{train}}(X, M))$.\footnote{The notation $\tau_{c, *}(\cdot)$ denotes the pushforward measure. We note that in our implementation $\tau_c$ is data dependent and we ignore this dependence to enable a simple analysis.} Divergences other than total-variation can be used, resulting in tighter bounds, e.g. see \citet{ben2010theory}. As we generate better counterfactuals this divergence decreases, and it can also be shown that $h^*$ and $h_{\text{aug}}^*$ coincide. Hence $\lambda_{\text{aug}}$ vanishes and the bound scales with $N^{-\frac{1}{2}}$, resulting in a gain of factor $d_{2,\text{train}}(Y, C)$
over the upper bound on $\widehat{\gR}^{\ell_{01}}_{\rvw}(h)$ in \Cref{eq:renyi_bound}. We discuss the details in the appendix, and in \Cref{sec:experiments} we show this empirically through simulations.

\textbf{Takeaways and additional baselines.} We emphasize that that the counterfactual datapoints should not be interpreted as ``more data'' in the sense of i.i.d training examples, they rather embody knowledge about how the causal mechanism that generates features $X$ acts under interventions on the attribute $C$ (as formalized in e.g. \citep{pearl2009causality, peters2017elements}). This translates into an improved sample complexity towards risk minimization on $P_\bot$. Counterfactuals are not the only type of causal knowledge that may be leveraged for learning more stable models. Many data dependent penalty terms have been proposed to impose conditional independence constraints drawn from the causal structure of the problem. Theory on these methods usually shows improved OOD performance under infinite data \citep{arjovsky2019invariant, wald2021calibration, puli2022outofdistribution, veitch2021counterfactual}. Our baselines include a method based on the Maximum-Mean Discrepency (MMD) from \citet{makar2022causally} who show improved sample complexity under a linear hypothesis class. 

\section{Experiments} \label{sec:experiments}

We empirically study the following questions:
(1) Can \ours enhance OOD performance of downstream classifiers? 
(2) Does it surpass the combination of reweighting and invariance penalties?
(3) Is it more effective than alternative augmentation techniques, thus demonstrating the usefulness of the causal graph?
(4) How sensitive is \ours to quality of counterfactuals?

These questions seek to establish causally-motivated augmentations as a practical approach for improving OOD performance. We address Q\#1,\#2 and \#3 through our theoretical foundation and across all empirical studies, while Q\#4 is explored in the synthetic experiments.
Further details about the experimental setup, including data statistics, model hyperparameters, and data splits, can be found in \Cref{app:exp_details}. \Cref{tab:tasks} provides an overview of the tasks we experiment with.

\begin{table}[h!]
    \centering
    \scalebox{0.8}{
    \begin{tabular}{p{2.8cm}|p{3cm}p{1.5cm}p{1.7cm}p{2.9cm}p{2.6cm}} \toprule
        Input ($x$) & Label ($y$) & ID Data & OOD Data & Spurious Feature ($c$) & \auxdata{} ($m$) \\ \midrule \midrule
        
        \multirow{3}{*}{\parbox{3cm}{Clinical Narratives}} & Condition Prediction &  \multirow{3}{*}{\parbox{2cm}{MIMIC-III }} & i2b2-2010 & \multirow{3}{*}{\parbox{2cm}{Caregiver ID}} & \multirow{3}{*}{\parbox{2cm}{Medications, Lab Results, Vitals}} \\
         & Note Segmentation & & partner data & & \\
         & Demographic Traits & & i2b2-2006 & & \\ \midrule

        Restaurant Reviews & Restaurant Rating & CEBaB & CeBAB-Spurious & Food-mention & Service, Noise, Ambiance, Food \\  \midrule
        Synthetic Data & $\{0, 1\}$ & \multicolumn{2}{c}{Gaussians} & $\{0, \cdots, 7\}$ &  -- \\ \bottomrule
    \end{tabular}
    }
    \caption{Description of all our tasks and their corresponding experimental setup.}
    \label{tab:tasks}
\end{table}

\textbf{Baselines.}
We compare \ours to several baselines:
\begin{itemize}[nolistsep]
    \item Observational - Baseline model trained on the original data. \textit{PubMED BERT} \cite{gu2021domain} for \textit{clinical narratives}, logistic regression for the \textit{restaurant reviews} and \textit{synthetic} experiments. \footnote{\Cref{app:exp_details} includes results where the Baseline model is also BioBERT, SentenceBERT or GPT3.}
    \item Reweighting - Baseline model with sample reweighting as in \citet{makar2022causally}.
    \item MMD - Baseline model with an MMD penalty as in \citet{veitch2021counterfactual, makar2022causally}.
    \item IRM - Baseline model with the IRMv1 penalty as in  \citet{arjovsky2019invariant}.
    \item GroupDRO - Baseline model trained with the GroupDRO objective as in \citet{sagawa2019distributionally}.
    \item Naive Augmentations - Baseline model on a dataset that also includes augmentations, generated by prompting an LLM to create more examples (without matching or diff-in-diff).
    \item Conditional Augmentations - Augmentations are generated by matching on \auxdata{} and prompting an LLM to create one example in the the style of the other.
\end{itemize}
The reweighting and MMD approaches are discussed and contrasted to counterfactual augmentation in \Cref{sec:cda}. IRM and GroupDRO are the most well-known principled methods for OOD generalization that are used in the literature. The augmentation approaches are compared here to demonstrate the importance of using the causal structure of the data.

\subsection{Clinical Narratives}

\textbf{Data.} We consider three representative clinical NLP tasks, \textit{clinical condition} prediction, \textit{note segmentation} and \textit{demographic traits} identification\footnote{See \Cref{app:exp_details} for results on the \textit{demographic traits} identification task.}, for which we have both ID and OOD data.
We utilize several electronic health records (EHR) datasets. We train on MIMIC-III \cite{johnson2016mimic}, a widely-used medical dataset containing over $2$ million notes from $38,597$ adult patients, $49,785$ hospital admissions, and $3,500$ healthcare professionals between 2001 and 2012. MIMIC-III is commonly used in NLP research for clinically-related tasks and for pre-training language models for the medical domain \cite{alsentzer2019publicly}. When available, we use i2b2 2006 and 2010 competitions as our held-out hospital dataset. In the note segmentation task, we use private held-out data.

\textbf{Generating notes from counterfactual caregivers.} To generate augmentations, we select caregivers with multiple patients and notes for more than one patient.
For each caregiver-patient pair where both their last progress note and discharge summary were written by that caregiver\footnote{During a patient's stay, progress notes capture its current state. When leaving the hospital, a discharge summary is written.}, we match them to similar patients having the same initial caregiver but a different one for their discharge summary.
In matching, we select patients with similar medications and lab results (denoted as patient's \auxdata{} $m$ in \Cref{tab:tasks}). We then generate counterfactual discharge summaries for matched patients using \Cref{alg:cdaug}(A) and train the model using original data and generated counterfactuals. 

\Cref{fig:results} presents results for \ours(A) using language model representations generated using these matched examples.  See \Cref{app:exp_details} for training details and results for \ours(A) with LLM prompts, and \Cref{app:note_examples} for synthetic note examples and the prompts used.

\definecolor{erm}{RGB}{102, 194, 165}
\definecolor{reweight}{RGB}{252, 141, 98}
\definecolor{aug1}{RGB}{166, 216, 84}
\definecolor{naive_aug}{RGB}{231, 138, 195}
\definecolor{mmd}{RGB}{141, 160, 203}
\definecolor{dro}{RGB}{253, 218, 13}
\definecolor{irm}{RGB}{210, 43, 43}


\begin{figure}[h!]
\begin{tikzpicture}[scale=0.8, line width=1pt]
\centering
\begin{axis}[
    ybar,
    ymin=60, ymax=93,
    enlarge x limits=0.4,
    legend style={at={(1.1,-0.15)}, anchor=north,legend columns=-1},
    ylabel={$F1$},
    symbolic x coords={ID (MIMIC-III), OOD (i2b2-2010)},
    xtick=data,
    title={\textit{(A) Clinical Conditions}}
    ]
\addplot+[color=black, fill=erm, error bars/.cd, y dir=both, y explicit] coordinates {(ID (MIMIC-III), 88.65) +-(0,0.5) (OOD (i2b2-2010), 64.6) +-(0,1)};
\addplot+[color=black, fill=reweight, error bars/.cd, y dir=both, y explicit] coordinates {(ID (MIMIC-III), 88.36) +-(0,1) (OOD (i2b2-2010), 67.26) +-(0,1)};
\addplot+[color=black, fill=mmd, error bars/.cd, y dir=both, y explicit] coordinates {(ID (MIMIC-III), 87.26) +-(0,1.2) (OOD (i2b2-2010), 67.38) +-(0,1.2)};
\addplot+[color=black, fill=irm, error bars/.cd, y dir=both, y explicit] coordinates {(ID (MIMIC-III), 88.79) +-(0,1.6) (OOD (i2b2-2010), 64.32) +-(0,2.0)};
\addplot+[color=black, fill=dro, error bars/.cd, y dir=both, y explicit] coordinates {(ID (MIMIC-III), 88.79) +-(0,0.6) (OOD (i2b2-2010), 66.70) +-(0,1.0)};
\addplot+[color=black, fill=naive_aug, error bars/.cd, y dir=both, y explicit] coordinates {(ID (MIMIC-III), 88.79) +-(0,0.4) (OOD (i2b2-2010), 66.75) +-(0,0.8)};
\addplot+[color=black, fill=aug1, error bars/.cd, y dir=both, y explicit] coordinates {(ID (MIMIC-III), 88.18) +-(0,0.5) (OOD (i2b2-2010), 72.76) +-(0,0.5)};
\legend{Observational, + Reweighting, ++ MMD, IRM, GroupDRO, Naive Aug., \ours(A)}
\end{axis}

\hspace{7cm}

\begin{axis}[
    ybar,
    ymin=60, ymax=93,
    enlarge x limits=0.4,
    ylabel={$F1$},
    symbolic x coords={ID (MIMIC-III), OOD (Private Held-Out)},
    xtick=data,
    title={\textit{(B) Note Segmentation}}
    ]
\addplot+[color=black, fill=erm, error bars/.cd, y dir=both, y explicit] coordinates {(ID (MIMIC-III), 91.14) +-(0,0.5) (OOD (Private Held-Out), 73.05) +-(0,1)};
\addplot+[color=black, fill=reweight, error bars/.cd, y dir=both, y explicit] coordinates {(ID (MIMIC-III), 90.09) +-(0,1) (OOD (Private Held-Out), 77.02) +-(0,1)};
\addplot+[color=black, fill=mmd, error bars/.cd, y dir=both, y explicit] coordinates {(ID (MIMIC-III), 89.14) +-(0,1.2) (OOD (Private Held-Out), 77.99) +-(0,1.2)};
\addplot+[color=black, fill=irm, error bars/.cd, y dir=both, y explicit] coordinates {(ID (MIMIC-III), 89.14) +-(0,1.8) (OOD (Private Held-Out), 75.80) +-(0,2.0)};
\addplot+[color=black, fill=dro, error bars/.cd, y dir=both, y explicit] coordinates {(ID (MIMIC-III), 89.14) +-(0,0.8) (OOD (Private Held-Out), 78.10) +-(0,1.0)};
\addplot+[color=black, fill=naive_aug, error bars/.cd, y dir=both, y explicit] coordinates {(ID (MIMIC-III), 91.72) +-(0,0.4) (OOD (Private Held-Out), 75.44) +-(0,0.8)};
\addplot+[color=black, fill=aug1, error bars/.cd, y dir=both, y explicit] coordinates {(ID (MIMIC-III), 91.85) +-(0,0.5) (OOD (Private Held-Out), 80.51) +-(0,0.5)};
\end{axis}
\end{tikzpicture}
\caption{Results ($F1$ averaged across 5 runs) for predicting \textit{clinical conditions} (A) and for clinical \textit{note segmentation} (B) from the text narratives. \ours(A) outperforms all baselines on OOD data.}
\label{fig:results}
\end{figure}
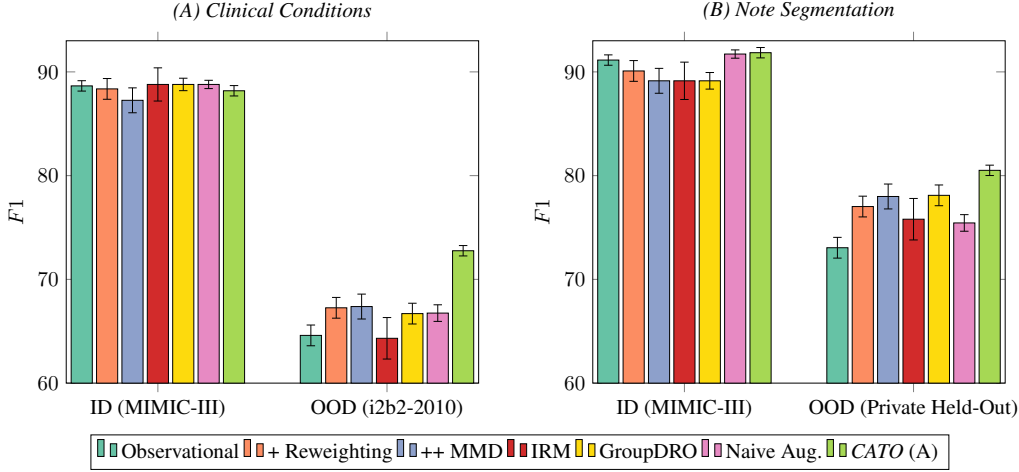

\textbf{Clinical Condition Prediction.}
\textit{Clinical condition} prediction is a concept extraction task focused on medical concepts in patient reports \cite{uzuner20112010}. Here we trained \textit{PubMED BERT} models on a subset of MIMIC-III, labelled using the same annotation guidelines as in i2b2-2010, the OOD dataset the models are tested on. As can be seen in the \Cref{fig:results}(A), in the ID setting only the naive augmentations improve performance slightly. In the OOD setting, all OOD methods help (\textit{reweighting}, \textit{MMD}, \textit{IRM}, \textit{GroupDRO}, \ours(A)), but our causally-motivated augmentation approach is substantially better than the alternatives. On average (across 5 runs), \ours(A) improves precision above the baseline by more than $7 \%$ (absolute), and recall by more than $8\%$. The naive augmentation approach improves over the vanilla \textit{PubMED BERT} model, but is outperformed by all OOD methods.

\textbf{Note Segmentation.}
In this task, models need to recognize sections in free-form clinical notes \cite{pomares2019current}. Given that section headers vary between hospitals, the models must discern sections based solely on the note content, excluding headers. As can be seen in \Cref{fig:results}(B), similarly to \textit{clinical condition} prediction, the diff-in-diff approach to augmentations (\ours(A)) substantially improved OOD performance, and as expected does not help ID. The naive augmentations are the best performing method ID, but is again outperformed by all other methods OOD.

\subsection{Restaurant Reviews}
\begin{wraptable}{r}{0.48\textwidth}
    \begin{tabular}{l|cc}
    
    Method & \textit{CeBAB} & \textit{CeBAB}-Spur. \\
    \midrule
    
    Observational & $\textbf{0.85}$ & $0.64$\\
    Reweighting & $0.84$ & $0.68$ \\
    Naive Aug. & $0.80$ & $0.62$ \\
    Conditional Aug. & $0.84$ & $0.70$ \\ 
     \ours(B) & $0.84$ & $\textbf{0.75}$ \\ \bottomrule
    \end{tabular}
    \caption{Accuracy on \textit{CeBAB} and \textit{CeBAB}-Spurious. \ours(B) outperforms all baselines when we introduce a spurious correlation.}\label{tb:restaurant}
    \vspace{-5mm}
\end{wraptable}

\textbf{Data.}
We use the \textit{CEBaB} dataset \cite{abraham2022cebab}, which consists of short restaurant reviews and ratings from \href{https://www.opentable.com/}{OpenTable}, including evaluations for food, service, noise, ambiance, and an overall rating. We used the train-exclusive split of the dataset, which contains $1,755$ examples. 
We construct two experimental settings: the original \textit{CeBAB} dataset, and a modified version, denoted as \textit{CeBAB}-Spurious, where there's a spurious correlation between training and deployment.

To construct \textit{CeBAB}-Spurious, we leverage the availability of both the original and perceived ratings for each review in \textit{CeBAB}. The original rating represents the reviewer's initial thoughts when writing the review, while the perceived rating indicates whether the review contains information about various restaurant attributes (e.g., food, service, noise, ambiance) and their associated sentiment. We utilize this unique data structure to capture reviewers' writing styles. Some reviewers are concise and provide limited descriptions, while others are more descriptive and include more information. To incorporate this variability, we introduce a new attribute called \textit{food-mention} to signify the presence of food-related information in a review. If the perceived food rating is either negative or positive, we assign a value of $1$ to the \textit{food-mention} attribute; otherwise, it is set to $0$. We subsample the data such that there is a correlation of $0.72$ between \textit{food-mention} and the outcome.

\textbf{Generating reviews with counterfactual food mentions.}
Following \Cref{alg:cdaug}, we generate counterfactual restaurant reviews conditional on food and overall ratings. We find matched examples for each review, select those with different food-mentions, and prompt an LLM to rewrite them, reflecting how the reviews would appear if the reviewer was more/less concise.

\textbf{Results.}
As shown in \Cref{tb:restaurant}, adding counterfactual augmentations leads to better OOD generalization, while naive data augmentation hurts model performance
In line with the sample complexity argument in \Cref{sec:cda}, 
conditional augmentation effectively doesn't add new data and therefore doesn't improve model performance.

\subsection{Synthetic Data}

To test sensitivity of \ours to quality of counterfactuals (Q\#4), we generate synthetic data for a binary classification problem where $K=8$ (cardinality of $C$). We sample $\tilde{P}(C \mid Y)$ to simulate varying degrees of spurious correlations. Then we draw $\rvx = [\rvx^*, \rvx_{\text{spu}}]$ from a Gaussian distribution,

\begin{align*}
    \rvx_i = \begin{bmatrix}
        \rvx_i^* \\
        \rvx_{\text{spu},i}
    \end{bmatrix} \sim \gN\left( 
    \begin{bmatrix}
        \vmu_{y_i} \\
        \vmu_{c_i}
    \end{bmatrix},
    \begin{bmatrix}
        \sigma^2 \mathbf{I}_{d^*} & 0 \\    
        0 & \sigma_{\text{spu}}^2\mathbf{I_{d_c}}
    \end{bmatrix}
    \right).
\end{align*}
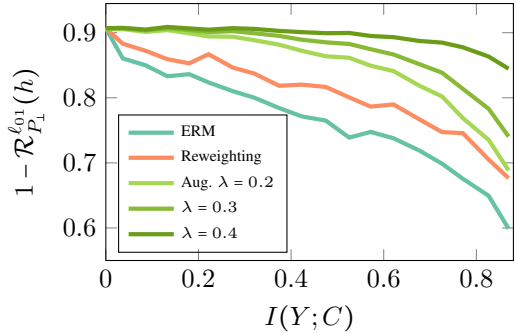
\begin{wrapfigure}{r}{0.5\textwidth}
  \begin{center}

\definecolor{erm}{RGB}{102, 194, 165}
\definecolor{reweight}{RGB}{252, 141, 98}
\definecolor{aug1}{RGB}{166, 216, 84}
\definecolor{aug2}{RGB}{136, 186, 54}
\definecolor{aug3}{RGB}{106, 156, 24}
\definecolor{mmd}{RGB}{141,160,203}

\begin{tikzpicture}
\begin{axis}[
    width=7cm,
    height=5cm,
    xlabel={$I(Y ; C)$},
    ylabel={$1-\gR^{\ell_{01}}_{P_{\bot}}(h)$},
    legend pos=south west,
    legend cell align=left,
    xmin=0, ymin=0.55, xmax=0.88,
    legend style={font=\tiny, fill=none}
]
\addplot[color=erm, ultra thick] coordinates {
    (0.0, 0.9068366666666666)
    (0.036530144915073226, 0.8605299999999999)
    (0.08587322600186534, 0.8499133333333333)
    (0.13290731615179355, 0.8330699999999999)
    (0.18005717938437266, 0.8362666666666666)
    (0.22183835797488063, 0.8235666666666667)
    (0.27383423521140027, 0.8102366666666666)
    (0.3197662740969753, 0.8003533333333335)
    (0.37439546431219417, 0.7841699999999999)
    (0.42319925888316573, 0.7714099999999999)
    (0.4749772565943993, 0.7649533333333334)
    (0.525561281547641, 0.7388200000000001)
    (0.5718405800161768, 0.7475999999999999)
    (0.6212477007995811, 0.7377900000000001)
    (0.6768078390858743, 0.7190399999999999)
    (0.7259363013203121, 0.6987766666666666)
    (0.7708129209486824, 0.6756466666666667)
    (0.8265925742470769, 0.6493866666666667)
    (0.8696064157042345, 0.5988733333333333)
};
    

\addplot[color=reweight, ultra thick] coordinates {
    (0.0, 0.9068366666666666)
    (0.036530144915073226, 0.8830966666666666)
    (0.08587322600186534, 0.8721733333333334)
    (0.13290731615179355, 0.85931)
    (0.18005717938437266, 0.8529633333333333)
    (0.22183835797488063, 0.8670066666666666)
    (0.27383423521140027, 0.8464666666666668)
    (0.3197662740969753, 0.8374600000000001)
    (0.37439546431219417, 0.8184633333333333)
    (0.42319925888316573, 0.8202533333333335)
    (0.4749772565943993, 0.8166333333333333)
    (0.525561281547641, 0.8006033333333333)
    (0.5718405800161768, 0.7864700000000001)
    (0.6212477007995811, 0.7897633333333333)
    (0.6768078390858743, 0.7669599999999999)
    (0.7259363013203121, 0.7473533333333333)
    (0.7708129209486824, 0.7456933333333331)
    (0.8265925742470769, 0.7050366666666666)
    (0.8696064157042345, 0.6764366666666668)
};

\addplot[color=aug1, ultra thick] coordinates {
    (0.0, 0.9059733333333334)
    (0.036530144915073226, 0.9055366666666667)
    (0.08587322600186534, 0.9013766666666666)
    (0.13290731615179355, 0.9040066666666667)
    (0.18005717938437266, 0.8990733333333333)
    (0.22183835797488063, 0.8942766666666667)
    (0.27383423521140027, 0.8938266666666667)
    (0.3197662740969753, 0.8886000000000001)
    (0.37439546431219417, 0.8815933333333335)
    (0.42319925888316573, 0.8721466666666667)
    (0.4749772565943993, 0.8638466666666665)
    (0.525561281547641, 0.8614766666666667)
    (0.5718405800161768, 0.8492533333333334)
    (0.6212477007995811, 0.8410400000000001)
    (0.6768078390858743, 0.8186266666666667)
    (0.7259363013203121, 0.8020499999999999)
    (0.7708129209486824, 0.7688466666666668)
    (0.8265925742470769, 0.7353999999999999)
    (0.8696064157042345, 0.6883399999999998)
};
\addplot[color=aug2, ultra thick] coordinates {
    (0.0, 0.9065333333333332)
    (0.036530144915073226, 0.9069300000000001)
    (0.08587322600186534, 0.9036133333333335)
    (0.13290731615179355, 0.9068600000000001)
    (0.18005717938437266, 0.9039766666666668)
    (0.22183835797488063, 0.9009699999999999)
    (0.27383423521140027, 0.9024466666666665)
    (0.3197662740969753, 0.8992166666666669)
    (0.37439546431219417, 0.8950866666666666)
    (0.42319925888316573, 0.8892233333333334)
    (0.4749772565943993, 0.8852633333333333)
    (0.525561281547641, 0.8826933333333332)
    (0.5718405800161768, 0.8746433333333333)
    (0.6212477007995811, 0.8664700000000001)
    (0.6768078390858743, 0.8517133333333331)
    (0.7259363013203121, 0.8385333333333335)
    (0.7708129209486824, 0.8138066666666666)
    (0.8265925742470769, 0.7830933333333333)
    (0.8696064157042345, 0.7405233333333333)
};
\addplot[color=aug3, ultra thick] coordinates {
    (0.0, 0.9069266666666667)
    (0.036530144915073226, 0.9074333333333334)
    (0.08587322600186534, 0.9048233333333333)
    (0.13290731615179355, 0.90888)
    (0.18005717938437266, 0.9065900000000001)
    (0.22183835797488063, 0.9046533333333333)
    (0.27383423521140027, 0.9069766666666669)
    (0.3197662740969753, 0.9058566666666668)
    (0.37439546431219417, 0.9027700000000001)
    (0.42319925888316573, 0.9008700000000001)
    (0.4749772565943993, 0.8992600000000001)
    (0.525561281547641, 0.8995966666666667)
    (0.5718405800161768, 0.8949866666666666)
    (0.6212477007995811, 0.8931433333333332)
    (0.6768078390858743, 0.8873599999999999)
    (0.7259363013203121, 0.8847666666666665)
    (0.7708129209486824, 0.8775833333333335)
    (0.8265925742470769, 0.8633766666666667)
    (0.8696064157042345, 0.8446033333333333)
};
    

\legend{ERM, Reweighting, Aug. $\lambda=0.2$, $\lambda=0.3$, $\lambda=0.4$}
\end{axis}
\end{tikzpicture} 
    \caption{OOD accuracy ($1-\gR^{l_{01}}_{P_{\bot}}(h)$) and $Y,C$ correlation strength ($I(Y ; C)$). \emph{Lower values} of $\lambda$ correspond to \emph{stronger corruptions} of the augmentations. Even with substantial corruption ($\lambda=0.2$) and strong correlation, augmentations outperform baselines.}\label{fig:synthetic}
  \end{center}
      \vspace{-5mm}
\end{wrapfigure}
In this case $\hat{\rvx}_i(c)$ is obtained by adding $\mu_{c} - \mu_{c_i}$ to $\rvx_{\text{spu},i}$. To corrupt our augmentation, we instead add $\xi_i\left(\mu_{c} - \mu_{c_i}\right)$ where $\xi_i$ is drawn from a truncated Gaussian centered at $\lambda\in{(0, 1)}$.
We train models with a fixed sample size (in the appendix we also examine varying sample sizes and additional types of corruption) and evaluate the trained models' accuracy on $P_{\bot}$ to examine the interplay between spurious correlation strength (measured by mutual information $I(Y; C)$), and counterfactual augmentation quality. As can be seen in \Cref{fig:synthetic}, corruptions degrade performance under stronger spurious correlations, though a strong corruption is required for reweighting to become preferable.

\section{Discussion}
\label{sec:dis}

In this work, we have presented a data augmentation approach based on the causal structure of \auxdata{} for improving OOD generalization, specifically focusing on text classification tasks. 
However, our approach is not without limitations. The validity of our assumptions, the specification of the causal graph and the quality of the counterfactual approximation all present challenges to address in future work. Further, our results suggest that performing data augmentation in an unprincipled manner can also hurt model performance. Utilizing additional techniques for OOD generalization, learning the causal structure directly from the data, and improving quality and reliability of the counterfactual approximation process can help mitigate these concerns.
Overall, we believe that causally-motivated data augmentation methods like ours can help address challenges in developing robust and reliable machine learning systems, particularly in safety-critical applications.

\clearpage
\bibliographystyle{unsrtnat}
\bibliography{bibliography}
\clearpage

\appendix
\section*{Appendix}

\section{Proofs of Formal Claims} \label{app:formal_things}
\textbf{Notation.} We will use random variables $C, Y, M, X$ with images $[K], \gY = [L], \gM, \gX$ respectively in our probabilistic causal models. For a function $\tau_c:\gX \times \gM \rightarrow \gX$, and measure $P$ over sets in $\gX\times \gM$, we denote by $\tau_{c, *}P(X, M)$ the pushforward measure \citep[\S 1.4]{tao2011introduction}. $\tau_c(\cdot)$ will be used to refer to the $c$-th coordinate of the output of a function $\tau:\gX \times \gM \rightarrow \gX^K$. The notation $\gH$ will be used for hypothesis classes where $h:\gX\rightarrow \gY$ for any $h\in{\gH}$. The $0-1$ loss $\ell_{01}:\gY\times\gY \rightarrow \{0, 1\}$ is given by $\ell_{01}(\hat{y}, y) = 1_{\hat{y} \neq y}$. For a node $V$ in a causal graph we will use $pa(V)$ for its causal parents.

For completeness we rewrite the definition of our data generating process from the main paper, this time adding the auxiliary data $M$ into our model.
\begin{repdefinition}{def:prob_setting}
Consider a probabilistic causal model with endogenous random variables $X,X^*,Y,C,M$ taking on values in $\gX,\gX^*,[L],[K],\gM$ and exogenous independent random variables \citep{peters2017elements} $N_X, N_{X^*, N_{Y}, N_C, N_M}$, where the induced graph is a DAG that satisfies the following,
\begin{itemize}
    \item $Y$ is $d$-separated from $X$ by $X^*, C, M$ and also by $X^*, C$.
    \item $Y, X^*$ are not descendants of $C$.
\end{itemize}
An anti-causal prediction problem with a spuriously-correlated attribute is a set of distributions $\gP$ obtained by all interventions on $C$ that replaces the distribution of exogenous noise $N_C$, mechanism $f_C(pa(C), N_C)$ with another mechanism (i.e. a measurable function $\tilde{f}(pa(C), N_C)$), or sets a fixed value (i.e. $do(C=c)$).
Under the settings of this problem, a learner is provided with a set $\left\{ (\rvx_i, y_i, c_i) \right\}_{i=1}^{N}$ sampled i.i.d from $P_{\text{train}}\in{\gP}$.

\end{repdefinition}
We denote by $P_{\bot}\in{\gP}$ the distribution obtained by intervening on $C$ and setting it to a uniform distribution, i.e. $P_{\bot}(X,X^*,Y,C,M) = K^{-1}\sum_{c\in{[K]}}{P(Y, X, X^*, M \mid do(C=c))}$. Note that the problem described by \cref{fig:dgp_purely_spu_ac} and \cref{def:prob_setting} of the main paper is a special case of this setting where $M$ is discarded, and $P_{\bot}$ coincides with setting $\tilde{P}(C \mid Y)$ to a uniform distribution.

Recall our assumption about perfect recovery of $X^*$.
\begin{assumption}
For an anti-causal prediction problem with a spuriously correlated attribute, we assume that $X^* = e(X)$ a.e. for some $e:\gX \rightarrow \gX^*$.
\end{assumption}
Under these conditions $h(\rvx) = \mathrm{arg}\max_{y\in{[L]}} P_{\bot}(Y=y \mid X=\rvx)$ is an optimal risk-invariant predictor as described below.
\begin{replemma}{lem:unconfounded_opt}
For the prediction problem in \cref{def:prob_setting}, the Bayes optimal classifier under the unconfounded distribution $P_\bot\in{\gP}$ where $C$ is uniformly distributed and independent of $Y$ is $h^*(\rvx) = \mathrm{arg}\max_{y\in{[K]}} P_\bot(Y=y \mid X^*=e(\rvx))$. It is a minimizer of $\min_{h:\gX\rightarrow [L] }\max_{P\in{\gP}}{\gR^{\ell_{01}}_{P}(h)}$ and $\gR^{\ell_{01}}_{P}(h^*)=\gR^{\ell_{01}}_{P_{\bot}}(h^*)$ for all $P\in{\gP}$.
\end{replemma}
\begin{proof}
    Assume $P_{\text{train}}\in{\gP}$ is the distribution from which our training data is obtained.
    We will show that any hypothesis satisfying $h(X)=g\circ e(X)$ for some $g:\gX^*\rightarrow \gY$ (i.e. that only depends on $X^*$) achieves the same risk over all $P\in{\gP}$. To this end note that for such a hypothesis we have,
    \begin{align*}
        R^{\ell_{01}}_{P_{\text{train}}}(h) &= \int{ \ell_{01}(h(X), Y)P_{\text{train}}(X \mid Y, C, X^*, M)P_{\text{train}}(Y, C, X^*, M) dX^*dXdYdCdM} \\ 
        &= \int{ \ell_{01}(g\circ e(X), Y)P_{\text{train}}(X \mid C, X^*, M)P_{\text{train}}(Y, C, X^*, M) dX^*dXdYdCdM} \\
        &= \int{ \ell_{01}(g(X^*), Y)P_{\text{train}}(X \mid C, X^*, M)P_{\text{train}}(Y, C, X^*, M) dX^*dXdYdCdM} \\ 
        &= \int{ \ell_{01}(g(X^*), Y)P_{\text{train}}(X^*, Y) dX^*dY} \\
        &= \int{ \ell_{01}(g(X^*), Y)P(X^*, Y) dX^*dY}.
    \end{align*}
    The first line writes down the expected risk explicitly, the second removes conditioning on $Y$ in the distribution on $X$ since we assumed $Y$ is $d$-separated from $X$ by $C, X^*, M$. In the third line we make it explicit that $h$ depends on $X^*$ alone, then we integrate out $X, C, M$. On the last line we remove the subscript $\text{train}$ to denote that this distribution in fixed across $P\in{\gP}$ as we assumed that $X^*,Y$ are non-descendants of $C$ (and members of $\gP$ are obtained by interventions on $C$).
    Now for any $\tilde{P}\in{\gP}$ we may repeat this derivation for $R^{l_{01}}_{\tilde{P}}(h)$ and we will obtain the same term (since $P(X^*, Y)$ are fixed regardless of the intervention applied in $P$, as we just argued), and we may conclude $R^{\ell_{01}}_{P_{\text{train}}}(h)=R^{\ell_{01}}_{\tilde{P}}(h)$.
    
    Next to show that the Bayes optimal classifier over $P_{\bot}$ is the min-max optimal classifier w.r.t $\gP$, consider the interventional distribution where $C$ is set to some fixed value $c\in{[K]}$, i.e. $P(X, X^*, Y \mid do(C=c))$. Under the graph we obtain from this intervention, $Y$ is $d$-separated from $X$ given $X^*$. Hence,
    \begin{align*}
    P(Y \mid X=\rvx, do(C=c)) &= \int_{X^*} P(Y \mid X^*, X=\rvx, do(C=c))P(X^* \mid X=\rvx, do(C=c))dX^* \\ 
    &= P(Y \mid X^*=e(\rvx), X=\rvx, do(C=c)) \\
    &= P(Y \mid X^*=e(\rvx), do(C=c)),
    \end{align*}
    where the first equality holds since $X^*=e(X)$ and the second from $d$-separation. Hence the Bayes optimal classifier under $P(Y, X \mid do(C=c))$ is $h^*(\rvx) = g\circ e(\rvx) = \mathrm{arg}\max_{y\in{[L]}}{P(Y=y \mid e(\rvx), do(C=c))}$. As argued earlier, since $Y, X^*$ are non-descendants of $C$, it holds that $P(Y \mid e(X), do(C=c))$ is fixed across all $c\in{[K]}$.
    Hence $h^*(\rvx)$ is the Bayes optimal classifier for all such interventional distributions and also for
    $P_{\bot}(X, Y) = \frac{1}{K}\sum_{c\in{[K]}}{P(X, Y \mid do(C=c))}$, and from our earlier discussion it is risk-invariant, i.e. $R^{\ell_{01}}_{P_{\bot}}(h^*) = R^{\ell_{01}}_{P}(h^*)$ for all $P\in{\gP}$, which also means $\max_{p\in{\gP}}{R^{\ell_{01}}_{P}(h^*)} = R^{\ell_{01}}_{P_{\bot}}(h^*)$. It is the min-max optimal classifier w.r.t $\gP$ since any $h \neq h^*$ will have $\max_{p\in{\gP}}{R^{\ell_{01}}_{P}(h)} \geq R^{\ell_{01}}_{P_{\bot}}(h) \geq R^{\ell_{01}}_{P_{\bot}}(h^*)$.
\end{proof}

Next we turn to prove a bound on sample complexity of counterfactual data augmentations.
\begin{replemma}{lem:generalization_bound}
Consider an anti-causal prediction problem with a spuriously-correlated attribute (\cref{def:prob_setting}), a measurable function $\tau:\gX\times \gM\rightarrow \gX^K$, and let $d_1(P, Q)$
denote the total variation distance between two distributions $P,Q$. Further let $h^*$ denote the optimal hypothesis w.r.t $\gR^{\ell_{01}}_{P_\bot}$ and let $\lambda_{\text{aug}} = \left[R_{\text{aug}}^{\ell_{01}}(h^*) + R_{P_\bot}^{\ell_{01}}(h^*) \right]$. For any hypothesis $h\in\gH$, and any $\delta\in{(0.5,1)}$ it holds that with probability at least $1-\delta$ over the draw of the training set,
\begin{align*}
    \gR^{\ell_{01}}_{P_\bot}(h) &\leq \widehat{\gR}^{\ell_{01}}_{\text{aug}}(h) + \sqrt{\frac{\log(1 / \delta)}{N}} + K^{-1}\cdot\sum_{c\in{[K]}}d_1\left(\tau_{c,*}\left( P_{\text{train}}(X, M) \right), P\left( X(c) \right) \right) + \lambda_{\text{aug}}.
\end{align*}
\end{replemma}
\begin{proof}
    Our first step is to show that for any hypothesis $h\in{\gH}$, if our augmentation process is exact in the sense that $\tau_c(X, M) = X(c)$ a.e., then the expected risk (i.e. risk taken over an infinitely large sample) on the augmented data coincides with that over the unconfounded distribution $P_{\bot}(X, Y) = P_{\text{unif}}(C)P(X, Y \mid do(C))$.
    \begin{align} \label{eq:population_equality}
        \gR_{\text{aug}}^{\ell_{01}}(h) &= \E_{P_{\text{train}}(C, Y, M, X)}{\left[  K^{-1} \sum_{c\in{[K]}}\ell_{01}(h\left(\tau_c(X, M)\right), Y) \right]} \nonumber \\
        &= K^{-1}\sum_{c\in{[K]}}{\E_{P_{\text{train}}(C, Y, M, X)}{\left[ {\ell_{01}(h\left(X(c)\right), Y)} \right]}} \nonumber \\
        &= K^{-1}\sum_{c\in{[K]}}{\E_{P_{\text{train}}(C, Y, X)}{\left[ {\ell_{01}(h\left(X(c)\right), Y(c))} \right]}} \nonumber \\
        &= K^{-1}\sum_{c\in{[K]}}{\E_{P(Y, X \mid do(C=c))}{\left[ {\ell_{01}(h\left( X \right), Y)} \right]}} \nonumber \\
         &= \gR_{P_{\bot}}^{\ell_{01}}(h).
    \end{align}
To bound $\gR^{\ell_{01}}_{\text{aug}}(h) - \hat{\gR}^{\ell_{01}}_{\text{aug}}(h)$ we note that $\{\rvx_i, y_i, \rvm_i\}_{i=1}^{N}$ are $i.i.d$ samples from a joint distribution, where we may consider the loss on each example as $K^{-1}\sum_{c\in{[K]}}{\ell_{01}(h(\tau_c(\rvx_i, \rvm_i), y_i))}$, then by standard results using the Hoeffding inequality, e.g. \citet[Corollary 2.11]{mohri2018foundations}, we get that for $\delta\in{(0.5, 1)}$,
\begin{align} \label{eq:single_hypo_hoeffding}
    \gR^{\ell_{01}}_{\text{aug}}(h) \leq \widehat{\gR}^{\ell_{01}}_{\text{aug}}(h) + \sqrt{\frac{\log(1/\delta)}{N}}.
\end{align}
Finally, to obtain our result consider any $c\in{[C]}$. Denote
\begin{align*} 
    \gR^{\ell_{01}}_{\text{aug}, c}(h) &:= \E_{P_{\text{train}}(Y, M, X)}{\left[\ell_{01}(h(\tau_c(X, M)) Y)\right]}, \\ \gR^{\ell_{01}}_{P_{\bot}, c}(h) &:= \E_{P(Y, X \mid do(C=c))}{\left[\ell_{01}(h(X), Y)\right]},
\end{align*}
and for $h^*$ denote $\gR^{\ell_{01}}_{\text{aug}, c}(h, h^*) := \E_{P_{\text{train}}(Y, M, X)}{\left[ \ell_{01}(h(\tau_c(X, M)), h^*(\tau_c(X, M))) \right]}$ and respectively for $\gR^{\ell_{01}}_{P_{\bot}}(h, h^*)$.
The rest of our derivation is along the lines of \citet[Theorem 2]{ben2010theory}. We use the distance
\begin{align*}
    d_{\gH\Delta\gH}(\tau_{c,*}P_{\text{train}}(X,M), P(X(c))) = 2\sup_{g\in{\gH\Delta\gH}} \left| P_{\text{train}}(g(\tau_c(X, M))=1) -  P(g(X(c))=1) \right|,
\end{align*}
where $\gH\Delta\gH = \{ g(\rvx) = 1_{h(\rvx)\neq h'(\rvx)} ~|~ h,h'\in{\gH} \}$ is a set of binary hypotheses, i.e. functions that mark disagreements between hypotheses in $\gH$. It is easy to see that $d_{\gH\Delta\gH}$ lower bounds $d_1$ which takes the supremum w.r.t all measurable subsets for the two measures, since the sets of inputs where $h(\rvx)=1$ are contained in those subsets. Also from \citep[Lemma 3]{ben2010theory} we have that for any hypotheses $h,h'\in{\gH}$ it holds that
\begin{align*}
    \left| R^{l_{01}}_{\text{aug}, c}(h,h') -  R^{l_{01}}_{P_{\bot}, c}(h,h') \right| \leq \frac{1}{2} d_{\gH\Delta\gH}\left(\tau_{c,*}P_{\text{train}}(X,M), P(X(c))\right)
\end{align*}
Then following the proof in \citet[Theorem 2]{ben2010theory}, where the first and third inequalities will rely on the triangle inequality for classification errors \citep{crammer2008learning}, we may get:
\begin{align*} \label{eq:bound_per_c}
\gR^{\ell_{01}}_{P_{\bot}, c}(h) &\leq \gR^{\ell_{01}}_{P_{\bot}, c}(h^*) + \gR^{\ell_{01}}_{P_{\bot}, c}(h, h^*) \\
& \leq \gR^{\ell_{01}}_{P_{\bot}, c}(h^*) + \gR^{\ell_{01}}_{\text{aug}, c}(h, h^*) + [\gR^{\ell_{01}}_{P_{\bot}, c}(h, h^*) - \gR^{\ell_{01}}_{\text{aug}, c}(h, h^*)] \\
& \leq \gR^{\ell_{01}}_{P_{\bot}, c}(h^*) + \gR^{\ell_{01}}_{\text{aug}, c}(h, h^*) + \frac{1}{2}d_{\gH\Delta\gH}\left(\tau_{c,*}P_{\text{train}}(X,M), P(X(c))\right) \\
& \leq \gR^{\ell_{01}}_{\text{aug}, c}(h) + \gR^{\ell_{01}}_{P_{\bot}, c}(h^*) + \gR^{\ell_{01}}_{\text{aug}, c}(h^*) + \frac{1}{2}d_{\gH\Delta\gH}\left(\tau_{c,*}P_{\text{train}}(X,M), P(X(c))\right) \\
& = \gR^{\ell_{01}}_{\text{aug}, c}(h) + \gR^{\ell_{01}}_{P_{\bot}, c}(h^*) + \gR^{\ell_{01}}_{\text{aug}, c}(h^*) + \frac{1}{2}d_{\gH\Delta\gH}\left(\tau_{c,*}P_{\text{train}}(X,M), P(X(c))\right)
\end{align*}
Finally, we note that $\gR^{\ell_{01}}_{P_{\bot}}(h) = K^{-1}\sum_{c\in{[K]}}{\gR^{\ell_{01}}_{P_{\bot}, c}(h)}$ and similarly we have that $\gR^{\ell_{01}}_{\text{aug}}(h) = K^{-1}\sum_{c\in{[K]}}{\gR^{\ell_{01}}_{\text{aug}, c}(h)}$, hence applying the above inequality for all $c\in{[K]}$ and averaging we get:
\begin{align*}
    \gR^{\ell_{01}}_{P_{\bot}}(h) &\leq \gR^{\ell_{01}}_{\text{aug}}(h) + \frac{1}{2}K^{-1}\sum_{c\in{[K]}}d_{\gH\Delta\gH}\left(\tau_{c,*}P_{\text{train}}(X,M), P(X(c))\right) + \lambda_{\text{aug}} \\
    &\leq \gR^{\ell_{01}}_{\text{aug}}(h) + K^{-1}\sum_{c\in{[K]}}d_1\left(\tau_{c,*}P_{\text{train}}(X,M), P(X(c))\right) + \lambda_{\text{aug}}.
\end{align*}
Combining with \cref{eq:single_hypo_hoeffding} we get the desired result.
\end{proof}

\subsection{Additional Causal Structures Where our Approach may be Used}
The problem setting we analyze in this work (see \cref{def:prob_setting}) captures a few interesting problems, mainly described as shortcut learning in the literature \citep{makar2022causally, puli2022nuisances, puli2023don}. However counterfactual data augmentation, and subsequently our approach of using \auxdata{} to perform it, are applicable to additional problem settings. \citet{wang2022unified} formalize domain-invariant learning under many data generating processes they refer to as Causally Invariant with Spurious Associations (CISA), where $Z$ (in our setting the caregiver $C$) is called the spurious factor of variation. These settings include a variety of causal and anti-causal prediction problems, and they assume that there exists some part of the input $X$, referred to as $X_Z^{\perp}$, that holds all the information in $X$ that is not caused by $Z$.
Whenever it holds that $Y \indep X \mid X_Z^\perp, Z $ the association between $Z$ and $Y$ is called ``purely spurious" and Thm.~ 9 in \citet{wang2022unified} states that for all such problems counterfactual data augmentation learns the optimal invariant predictor over the training distribution. Hence in all such settings, improving counterfactual data augmentation with \ours{} can be beneficial towards OOD generalization. We refer the interested reader to \citep{wang2022unified} for further details on CISA problems and their properties.

\begin{figure}
        \centering 
        \raisebox{0pt}[\dimexpr\height-0.6\baselineskip\relax]{
        \includegraphics[scale=0.6]{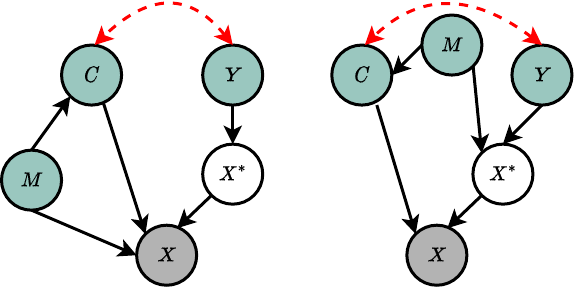}}%
        \caption{Possible causal structures that involve the auxiliary data $M$, where unobserved $M$ corresponds to unobserved confounding between $X$ and $C$.}
        \label{fig:causal_structs_M}
\end{figure}

We further note that in our work we excluded the auxiliary data $M$ from the causal model as we are agnostic to its specific causal relation with other factors in the data, so long as it satisfies \cref{ass:strong_ignorability} of strong ignorability. \cref{fig:causal_structs_M} depicts two potential structures that may adhere to this assumption.

\section{Experimental Details} \label{app:exp_details}

We provide here further details about the experimental setup, the datasets we use, hyperparameters chosen for training the models, and data splits. We also include additional experiments that were omitted from the main paper for brevity, including experiments on identifying \textit{demographic traits} in clinical narratives. 

\subsection{Clinical Narratives}

\subsubsection{Data}
We describe here the \textit{MIMIC-III} \textit{i2b2-2006} and \textit{i2b2-2010} datasets.

\paragraph{MIMIC-III.}

The \textit{MIMIC-III} (Medical Information Mart for Intensive Care III) dataset is a large, publicly available database containing detailed and anonymized health-related data associated with over 40,000 patients who stayed in critical care units at the Beth Israel Deaconess Medical Center in Boston, Massachusetts between 2001 and 2012.
\textit{MIMIC-III} is a rich resource for researchers in various fields, such as medicine, data science, artificial intelligence, and healthcare analytics. The dataset contains a diverse range of data types, including demographics, vital signs, laboratory test results, medications, and clinical notes. The dataset contains over $2$ million clinical notes contributed by over $3,500$ distinct healthcare professionals, including doctors, nurses, and other clinicians, with an average of $571$ notes per author. 

The notes in the \textit{MIMIC-III} dataset come in various types, reflecting the diverse aspects of patient care and documentation in the intensive care setting. Some of the most common note types include:

\begin{itemize}
    \item Nursing/Progress notes: These are daily notes written by nurses or other care providers, documenting the patient's progress, condition, and care provided.
    \item Radiology reports: Reports written by radiologists after interpreting medical imaging studies (e.g., X-rays, MRIs, CT scans).
    \item ECG reports: Reports documenting the interpretation of electrocardiogram results.
    \item Discharge summaries: Comprehensive summaries written by physicians when a patient is discharged from the hospital, outlining the patient's hospital course, treatments, and follow-up instructions.
    \item Physician consult notes: Notes written by specialists when consulted by the primary care team to provide their expert opinion on specific medical issues.
    \item Pharmacy notes: Notes documenting medication-related information, including dosing, administration, and potential drug interactions.
    \item Social work notes: Notes related to the patient's psychosocial status, including social and family support, living arrangements, and other relevant factors.    
\end{itemize}

\paragraph{i2b2-2006.}
The i2b2 (Informatics for Integrating Biology and the Bedside) initiative is a collaborative effort that aims to develop new methods and tools for biomedical research. It focuses on the development of a scalable computational infrastructure that can be used to accelerate the translation of basic research findings into clinical applications. As part of this effort, i2b2 has hosted several shared tasks and challenges related to natural language processing and machine learning in healthcare.

In 2006, the first i2b2 challenge, known as the \textit{i2b2-2006} challenge, was conducted, focusing on the identification of obesity and its comorbidities in discharge summaries. The dataset provided for the challenge contained $694$ de-identified discharge summaries, which were randomly selected from the Research Patient Data Registry (RPDR) at Partners HealthCare. The dataset was divided into a training set of $514$ discharge summaries and a test set of $180$ discharge summaries.
It is important to mention that the \textit{i2b2-2006} dataset is relatively small compared to the \textit{MIMIC-III} dataset and does not provide detailed information about the number of distinct authors or the average number of notes per author.

However, the discharge summaries typically include various sections such as patient demographics, admission and discharge dates, admission diagnoses, hospital course, procedures, medications, and follow-up plans. These summaries are generally written by physicians at the time of patient discharge, providing an overview of the patient's medical condition, treatment received, and overall hospital stay.

\paragraph{i2b2-2010.}
The \textit{i2b2-2010} challenge, also known as the i2b2/VA challenge, was a shared task organized by the i2b2 (Informatics for Integrating Biology and the Bedside) initiative in collaboration with the US Department of Veterans Affairs (VA). The challenge aimed to encourage the development of natural language processing (NLP) and machine learning techniques for extracting medical concepts from clinical narratives. Specifically, the \textit{i2b2-2010} challenge focused on the identification of medical problems, tests, and treatments from free-text clinical records.

The dataset provided for the \textit{i2b2-2010} challenge contained $826$ de-identified clinical records, which were sourced from three different institutions: Partners HealthCare, the University of Pittsburgh Medical Center (UPMC), and the VA. The dataset was divided into a training set of $349$ records and a test set of $477$ records.

Similar to the \textit{i2b2-2006} challenge, the \textit{i2b2-2010} dataset is relatively small compared to the \textit{MIMIC-III} dataset and does not provide detailed information about the number of distinct authors or the average number of notes per author. The clinical records in the dataset are composed of diverse note types, such as discharge summaries, progress notes, radiology reports, and pathology reports, contributed by physicians, nurses, and other healthcare professionals.

While the dataset does not provide specific information about the number of distinct authors, the fact that the notes were contributed by different types of healthcare professionals across multiple institutions increases the dataset's diversity, making it more representative of real-world clinical settings.

\subsubsection{PubMED BERT}

In our clinical narratives experiments, we use \textit{PubMED BERT} \cite{gu2021domain}, a variant of of the original BERT model \cite{devlin2018bert}, as our vanilla model. That is, all of the baselines and \ours all use it either for embedding clinical text or for predicting \textit{conditions}, \textit{demographic traits} and \textit{note segments}.

\textit{PubMED BERT} is a BERT-based (Bidirectional Encoder Representations from Transformers) model that has been pre-trained specifically on biomedical and scientific text data \cite{gu2021domain}. The model leverages the BERT architecture, which is a transformer-based deep learning model that has gained significant attention in natural language processing (NLP) for its state-of-the-art performance across a wide range of tasks.

\textit{PubMED BERT} is pre-trained on a large corpus of approximately $14$ million biomedical abstracts from the PubMed database, which is a comprehensive repository of biomedical literature. By pre-training the model on domain-specific data, \textit{PubMED BERT} is expected to have a better understanding of biomedical concepts, terminology, and language patterns compared to general domain models like BERT-base and BERT-large \cite{devlin2018bert}.

The main advantage of using \textit{PubMED BERT} for biomedical text mining tasks is its domain-specific knowledge, which can lead to improved performance and more accurate results when fine-tuned on various downstream tasks, such as named entity recognition, relation extraction, document classification, and question answering. Since \textit{PubMED BERT} is pre-trained on a large corpus of biomedical text, it is better suited to capturing the unique language patterns, complex terminology, and the relationships between entities in the biomedical domain.

\paragraph{Hyperparameters for Fine-Tuning PubMED BERT on \textit{MIMIC-III}.}

In our study, we leveraged a pre-trained \textit{PubMED BERT} model and fine-tuned it on the \textit{MIMIC-III} dataset. During pre-training, the model employed masked language modeling and next sentence prediction objectives. The architecture consisted of $12$ layers, $768$ hidden units, and $12$ attention heads. For task-specific optimization, we used the following hyperparameters: a $3e-5$ learning rate with a linear warmup during the initial $10\%$ of training steps, a batch size of $32$, a maximum sequence length of $512$ tokens, and a dropout rate of $0.1$. The AdamW optimizer was applied with a $0.01$ weight decay and a $1.0$ gradient clipping threshold. To prevent overfitting, early stopping was based on validation loss and used a $3$-epoch patience. The fine-tuning process ran for up to $20$ epochs, unless early stopping criteria were met sooner. 

The fine-tuning process was executed on a high-performance computing cluster with multiple NVIDIA Tesla V100 GPUs, each equipped with $32$ GB of memory, using the \textit{PyTorch} deep learning framework \cite{NEURIPS2019_9015}. The dataset was preprocessed and tokenized using the \textit{HuggingFace Transformers} library \cite{wolf2019huggingface}.

\subsubsection{\textit{Demographic Traits} Detection}

\textit{Demographic Traits} detection is the task of identifying residual private information in the clinical note, after removing the known identifier types (names, ages, dates, addresses, ID's, etc.) \cite{feder2020active}. We train all models on a subset of \textit{MIMIC-III} and test on \textit{i2b2-2006}. \Cref{tab:results-dem} presents our results. While performance gains from the Causal Augmentation approach are not as large as in the other clinical NLP tasks, its is still the best method in terms of $F1$ score on out-of-distribution examples.

\begin{table*}[h!]
    \centering
    \small
    \begin{tabular}{l|ccc|ccc}
     & \multicolumn{3}{c}{ID (\textit{MIMIC-III})} &  \multicolumn{3}{|c}{OOD (\textit{i2b2-2006})} 
 \\ \toprule
     & P & R & F1 & P & R & F1  \\ \midrule \midrule  
    \textit{PubMED BERT} & 80.61 & 78.12 & 79.34 & 53.32 & 90.1 & 66.92 \\
    + \textit{Re-Weighting} & 81.31 & 78.57 & \textbf{79.92} & 56.75 & 91.38 & 70.02 \\
    ++ \textit{MMD} & 80.68 & 78.84 & 79.75 & 56.19 & \textbf{91.49} & 69.62 \\
    \textit{Bio BERT} & 79.5 & 77.63 & 78.55 & 53.32 & 89.84 & 66.71 \\
    \textit{Sentence BERT} & 79.29 & 76.18 & 76.53 & 52.22 & 89.82 & 65.04 \\
    \textit{GPT3} & 78.31 & 76.01 & 77.18 & 52.73 & 88.52 & 63.98 \\
    \textit{Naive Aug.}  & \textbf{81.45} & \textbf{79.35} & 80.39 & 52.9 & 89.58 & 66.52 \\
    \textbf{\textit{Causal Aug.}} & 80.65 & 78.84 & 79.73 & \textbf{59.76} & 90.16 & \textbf{71.88}\\ \bottomrule
    \end{tabular}
    \caption{Results (averaged across 5 runs) for predicting demographic traits from the text narratives on in-distribution and out-of-distribution data.}
    \label{tab:results-dem}
\end{table*}

\subsection{Restaurant Reviews}

\paragraph{Data.}
We use the \textit{CEBaB} dataset \cite{abraham2022cebab}, which consists of short restaurant reviews and ratings from \href{https://www.opentable.com/}{OpenTable}, including evaluations for food, service, noise, ambiance, and an overall rating. For our experiments, we used the train-exclusive split of the dataset, which contains $1,755$ examples. 

To analyze the data, we transformed the overall rating into a binary outcome. The original rating scale ranges from $1$ to $5$, and we classified a rating of $3$ or higher as $1$, and anything below as $0$. We utilized a bag-of-words model with \textit{CountVectorizer} and fitted logistic regression models from the \textit{sklearn} library \cite{pedregosa2011scikit}.



To investigate these questions, 
we construct two experimental settings: the original \textit{CeBAB} dataset, and a modified version, denoted as \textit{CeBAB}-Spurious, where there's a spurious correlation between training and deployment.

The data is randomly split into a training set with $1,000$ examples and a test set with $755$ examples.
We explore two data augmentation schemes:
\begin{enumerate}
    \item Naive data augmentation: This approach involves randomly selecting two reviews from the dataset and prompting \textit{GPT-4} \cite{openai2023gpt4}  to rewrite one restaurant review in the style of the other. By applying the naive augmentation, we obtain an additional $1,000$ training examples.
    \item Conditional data augmentation : We match the ratings and sub-ratings in the reviews to create pairs. We then prompt \textit{GPT-4} to rewrite one review to match the style of the other. Because not all pairs have matches in this case, the conditional data augmentation generates $926$ augmentations. See \Cref{app:exp_details} for details of the prompt.
\end{enumerate}

\paragraph{Generating reviews with counterfactual food mentions.}
Following the counterfactual generation procedure in \Cref{alg:cdaug}, we generate counterfactual restaurant reviews conditional on food rating and overall rating. For each review, we first find a set of matched examples. We then select the subset that has different food-mention attribute and prompt \textit{GPT-4} to rewrite.
This results in $2,537$ augmentations. 
The counterfactual augmentation should capture what the reviews should look like had a reviewer been more/less concise.  
Following \Cref{alg:cdaug}, we generate counterfactual restaurant reviews conditional on food and overall ratings. We find matched examples for each review, select those with different food-mentions, and prompt a \textit{GPT-4} to rewrite them, reflecting how the reviews would appear if the reviewer was more/less concise.

\paragraph{Prompt Example. }

\begin{minted}{python}
helper_prompt = """
you are a very helpful, diligent, and intelligent language model assistant, 
your task to generate counterfactual restaurant reviews, 
that is what the restaurant review would be if it is given a different rating.
You will be given an original restaurant review and a comparator review
Your task is to rewrite the original review, such that it will have the same 
review score as the comparator review. 
The rating is with respect to ambiance, food, noise, and service.
---- EXAMPLE INPUT - START -----

original_review: [],
original_ratings: [
rating_ambiance: score,
rating_food: score,
rating_noise: score,
rating_service: score
]

compare_reviews:[]
compare_ratings:[
rating_ambiance: score,
rating_food: score,
rating_noise: score,
rating_service: score
]


---- EXAMPLE INPUT - END -----
ANSWER FORMAT: 
{
original_review: [],
original_score: [],
rewrite_review: [],
}

"""
\end{minted}

\subsection{Synthetic Data}
As described in the main paper we study a binary classification problem where $K=8$ (cardinality of $C$), and sample $\tilde{P}(C \mid Y)$ to simulate varying degrees of the spurious correlation (specifically, we draw ). Then we draw $\rvx = [\rvx^*, \rvx_{\text{spu}}]$ from a Gaussian distribution,
\begin{align*}
    \rvx_i = \begin{bmatrix}
        \rvx_i^* \\
        \rvx_{\text{spu},i}
    \end{bmatrix} \sim \gN\left( 
    \begin{bmatrix}
        \vmu_{y_i} \\
        \vmu_{c_i}
    \end{bmatrix},
    \begin{bmatrix}
        \sigma^2 \mathbf{I}_{d^*} & 0 \\    
        0 & \sigma_{\text{spu}}^2\mathbf{I_{d_c}}
    \end{bmatrix}
    \right).
\end{align*}
In our simulations, we set $d^*=10, d_{\text{spu}}=300$ and $\sigma^2_{spu}=0.05, \sigma=0.01d^*$ to make the max-margin classifiers depend on the spurious features. The parameters $\mu_{y_i}, \mu_{c_i}$ are drawn uniformly from a sphere of norm $1/3$ and $60$, respectively. For the corruptions of augmentations where we add $\xi_i(\mu_c - \mu_{c_i})$, the $\xi_i$ variables are drawn from a truncated Gaussian centered at $\lambda$ with standard deviation $0.1$.

\begin{figure}[!h]
    \centering

\definecolor{erm50}{RGB}{102, 194, 165}
\definecolor{reweight}{RGB}{252, 141, 98}
\definecolor{aug1}{RGB}{166, 216, 84}
\definecolor{aug2}{RGB}{136, 186, 54}
\definecolor{aug3}{RGB}{106, 156, 24}
\definecolor{erm200}{RGB}{141,160,203}

\definecolor{reweight}{RGB}{252, 141, 98}
\definecolor{aug1}{RGB}{166, 216, 84}
\definecolor{aug2}{RGB}{136, 186, 54}
\definecolor{aug3}{RGB}{106, 156, 24}
\definecolor{erm200}{RGB}{141,160,203}

\begin{tikzpicture}
\begin{axis}[
    width=7cm,
    height=5cm,
    xlabel={$N$},
    ylabel={$1-\gR^{\ell_{01}}_{P_{\bot}}(h)$},
    legend pos=south east,
    legend cell align=left,
    xmin=0, ymin=0.55, xmax=2000,
    legend style={font=\tiny, fill=none}
]
\addplot[color=erm, ultra thick] coordinates {
    (50, 0.5222111111111111)
    (100, 0.5510777777777777)
    (200, 0.6035277777777778)
    (400, 0.6688055555555555)
    (700, 0.6975777777777777)
    (1100, 0.7179888888888889)
    (1600, 0.7389055555555556)
    (1999, 0.7465833333333333)
};
    

\addplot[color=reweight, ultra thick] coordinates {
    (50, 0.5514222222222223)
    (100, 0.6090777777777777)
    (200, 0.6561666666666668)
    (400, 0.7158166666666667)
    (700, 0.7668666666666668)
    (1100, 0.7827722222222223)
    (1600, 0.8188055555555557)
    (1999, 0.8340555555555554)
};

\addplot[color=aug1, ultra thick] coordinates {
    (50, 0.6053777777777777)
    (100, 0.6581555555555555)
    (200, 0.7178388888888889)
    (400, 0.7677777777777777)
    (700, 0.7931222222222224)
    (1100, 0.8063944444444444)
    (1600, 0.8175277777777779)
    (1999, 0.8219166666666667)
};
\addplot[color=aug2, ultra thick] coordinates {
    (50, 0.6449944444444444)
    (100, 0.7210777777777777)
    (200, 0.7711333333333333)
    (400, 0.8100222222222222)
    (700, 0.8248833333333333)
    (1100, 0.8338388888888889)
    (1600, 0.8436833333333333)
    (1999, 0.8489333333333333)
};

\legend{ERM, Reweighting, Aug. $\lambda=0.2$, $\lambda=0.3$}
\end{axis}
\end{tikzpicture}
    \caption{OOD accuracy ($1-\gR^{l_{01}}_{P_{\bot}}(h)$) for growing size of i.i.d training set $N$. We run $15$ repetitions where $\tilde{P}(C \mid Y)$ are drawn randomly with correlation strength $I(Y ; C) = 0.743 \pm 0.019$. With large amounts of data, the reweighting method approaches optimal performance and may outperform solutions based on corrupted data augmentation (e.g. it surpasses the more heavily corrupted data augmentation with $\lambda=0.2$).}\label{fig:synthetic2}
\end{figure}
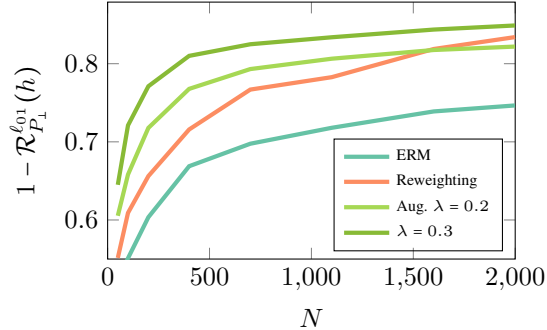

For the results in \cref{fig:synthetic} of the main paper we set the number of training examples $N$ at $600$ and the distributions $\tilde{P}(C \mid Y)$ are sampled such that for each interval of size $0.05$ between $0$ and $0.9$ for the values of $I(Y; C)$, we draw $30$ instances within that interval. In \cref{fig:synthetic2} we give results for another experiment where we plot curves for reweighting, ERM and corrupted augmentation under several values of $N$ under a strong spurious correlation. We draw values for $\tilde{P}(C \mid Y)$ such that that $I(Y;C)$ is in $[0.7, 0.8]$ (mean $0.743$ and standard deviation $0.019$ with $15$ repetitions). Considering the bounds in \cref{eq:renyi_bound} and the one in \cref{lem:generalization_bound}, we expect that as $N$ grows the reweighting method will approach optimal accuracy, while the augmentation methods will incur some bias due to the imperfect approximation of counterfactuals. In \cref{fig:synthetic2} we see that the augmentation based methods begin to plateau, while the reweighting method gains larger improvement from the growing $N$, and outperforms the more heavily corrupted augmentation method ($\lambda=0.2$) with a large enough training set. The best possible mean accuracy in this simulation is $0.934$ (as we gather from training a classifier from $\rvx^*$ alone), the best accuracy obtained by the methods in the figure is $0.849$ (Aug. $\lambda=0.3$ with $N=1999$).

\section{Synthetic Note Examples} \label{app:note_examples}

While it is not possible to share real notes from the \textit{MIMIC-III} or any of the \textit{i2b2} datasets, we can share how a typical note might look like by generating synthetic notes with LLMs. Here, we generate a synthetic note using \textit{GPT-4,} and then generate a counterfactual note using our prompt:

\noindent\fbox{\begin{minipage}{\textwidth}
\textbf{Discharge Summary Note}

\textbf{Dr. Jane Doe}

\textbf{March 29, 2023}

\textbf{Patient Name:} John Doe\

\textbf{Admission Date:} March 25, 2023\

\textbf{Discharge Date:} March 29, 2023\

\textbf{Diagnosis:} Pneumonia\

\textbf{History of Present Illness:}
Mr. Doe is a 58-year-old male with a past medical history of hypertension who presented to the emergency department with a 3-day history of fever, cough, and shortness of breath. Chest X-ray showed bilateral infiltrates, and he was admitted for treatment of community-acquired pneumonia.

\textbf{Hospital Course:}
The patient was started on intravenous antibiotics and supplemental oxygen. Over the course of his hospitalization, his oxygen requirements decreased, and his fever and cough resolved. He was able to tolerate a regular diet and was mobilized with physical therapy.

\textbf{Consultations:}
Infectious disease was consulted, and the patient's antibiotics were adjusted based on culture results.

\textbf{Discharge Medications:}
\begin{enumerate}
\item Levofloxacin 750mg daily for 10 days
\item Acetaminophen 650mg every 6 hours as needed for fever or pain
\item Albuterol inhaler 2 puffs every 4 hours as needed for shortness of breath
\end{enumerate}

\textbf{Follow-up:}
The patient was instructed to follow up with his primary care physician in 1 week for a repeat chest X-ray and evaluation of his symptoms.

\textbf{Discharge Instructions:}
\begin{enumerate}
\item Finish the entire course of antibiotics as prescribed.
\item Use the albuterol inhaler as needed for shortness of breath.
\item Rest and avoid strenuous activity until cleared by your doctor.
\item Follow up with your primary care physician in 1 week.
\end{enumerate}

\textbf{Discharge Condition:}
The patient was discharged in stable condition with improved symptoms.

\end{minipage}}

\newpage
Here's a counterfactual discharge summary note written by a doctor with a different writing style:

\noindent\fbox{\begin{minipage}{\textwidth}

\textbf{Patient Name:} John Doe\

\textbf{Admission Date:} March 25, 2023\

\textbf{Discharge Date:} March 29, 2023\

\textbf{Diagnosis:} Pneumonia\

\textbf{HPI:}
John is a middle-aged man who came in with a fever and cough that he's had for three days. He was diagnosed with pneumonia and admitted.

\textbf{Hospital Course:}
John was given antibiotics and oxygen. He slowly got better and was able to eat normally and move around more. He saw an infectious disease specialist who adjusted his treatment.

\textbf{Consultations:}
Infectious disease saw John and changed his medicine.

\textbf{Discharge Medications:}
\begin{enumerate}
\item Levofloxacin 750mg once a day for 10 days
\item Acetaminophen 650mg every 6 hours as needed for fever or pain
\item Albuterol inhaler 2 puffs every 4 hours as needed for shortness of breath
\end{enumerate}

\textbf{Follow-up:}
Follow up with PCP in 1 week.

\textbf{Discharge Instructions:}
\begin{enumerate}
\item Finish your antibiotics.
\item Use the inhaler if you need it.
\item Rest and avoid heavy activity until you feel better.
\item Follow up with your doctor next week.
\end{enumerate}

\textbf{Discharge Condition:}
Stable, going home.
\end{minipage}}

As can be seen from these examples, the counterfactual note is much more concise and to-the-point than the original example. The language used is more direct and less descriptive, and there is less detail provided about the patient's course of treatment.

\section{Possible Limitations of LLMs in Generating Augmented Datasets}
As mentioned in our discussion, there are several possible limitations that should be carefully considered before applying our approach in practice, especially in high-stakes applications such has medical notes classification. We list some of the main possible limitations and points to consider, along with a short discussion on each.
\begin{itemize}
\item \emph{LLM generation quality}: LLMs vary in their ability to generate realistic text. It is possible that LLMs introduce biases into our problem, inherited from their own training data. This requires further study, however from our manual examination we found their quality satisfactory (see \cref{app:note_examples} for generation examples) and that OOD generalization also improved for models trained on the augmented data they generate. We also include experiments with several types of LLMs in \cref{app:exp_details} to verify that our findings are consistent across the types of LLMs we considered.
\item \emph{Counterfactual approximation}: Other than generation quality, the additional challenge in using LLMs for counterfactual data augmentation is our ability to elicit a good approximation to the counterfactual text. Our methods rely on principles from causal inference to advance disciplined approaches for this task. While further studies are required (e.g. systematically comparing small sets of manual re-writes of texts to the elicited LLM output), we view our work as a promising first step in this direction, which we expect to be significantly extended and improved in future work.
\item \emph{Effect of biases on OOD generalization}: Since we focus on OOD generalization, the limitations and possible biases mentioned above must be weighed within this context. Namely, we should bear in mind that even though generation may be biased, this bias is only harmful when it affects the generalization of a downstream classifier, and this is what we evaluate. Further, in OOD generalization we consider cases where the training data is biased in the first place, and training a standard predictive model also results in a biased solution. Hence we must weigh risks and limitations of alternative solutions vs. those of LLMs.
\end{itemize}

\end{document}